\documentclass[times, review, 10pt]{elsarticle}

\usepackage{amssymb}
\usepackage{amsmath}
\usepackage{wrapfig}
\usepackage{makecell}

\begin{document}

\begin{frontmatter}



\title{Text-to-image Diffusion Models in  Generative AI:  A Survey} %% Article title


\author{Chenshuang~Zhang$^a$
Chaoning Zhang$^b$\footnote{Corresponding author.
}, Mengchun Zhang$^a$, \\
In So Kweon$^a$, Junmo Kim$^a$} %% Author name

    % ^{1}$\and
    % $^{2}$\and
    % $^{2}$\and
    % $^{1}$\and
    % $^{1}$\
    % \affiliations
    % $^{1}$Korea Advanced Institute of Science and Technology, South Korea\\
    % $^{2}$Kyung Hee University, South Korea\\
    % \emails
    % zcs15@kaist.ac.kr, 
    % chaoningzhang1990@gmail.com, 
    % iskweon77@kaist.ac.kr


\affiliation{organization={Korea Advanced Institute of Science and Technology (KAIST)},%Department and Organization
            % addressline={123}, 
            % city={123},
            % postcode={123}, 
            % state={123},
            % country={South Korea}
            }

\affiliation{organization={Kyung Hee University},%Department and Organization
            % addressline={123}, 
            % city={123},
            % postcode={123}, 
            % state={123},
            % country={South Korea}
            }

%% Abstract
\begin{abstract}
This survey reviews the progress of diffusion models in generating images from text, ~\textit{i.e.} text-to-image diffusion models. As a self-contained work, this survey starts with a brief introduction of how diffusion models work for image synthesis, followed by the background for text-conditioned image synthesis. Based on that,  we present an organized review of pioneering methods and their improvements on text-to-image generation. We further summarize applications beyond image generation, such as text-guided generation for various modalities like videos, and text-guided image editing. Beyond the progress made so far, we discuss existing challenges and promising future directions.

% Abstract text.
\end{abstract}

%% Keywords

\end{frontmatter}


\begin{figure}[t!]
    \centering
    \setlength{\tabcolsep}{0.0pt}
    \begin{tabular}{ccc}
    \includegraphics[width=0.31\textwidth]{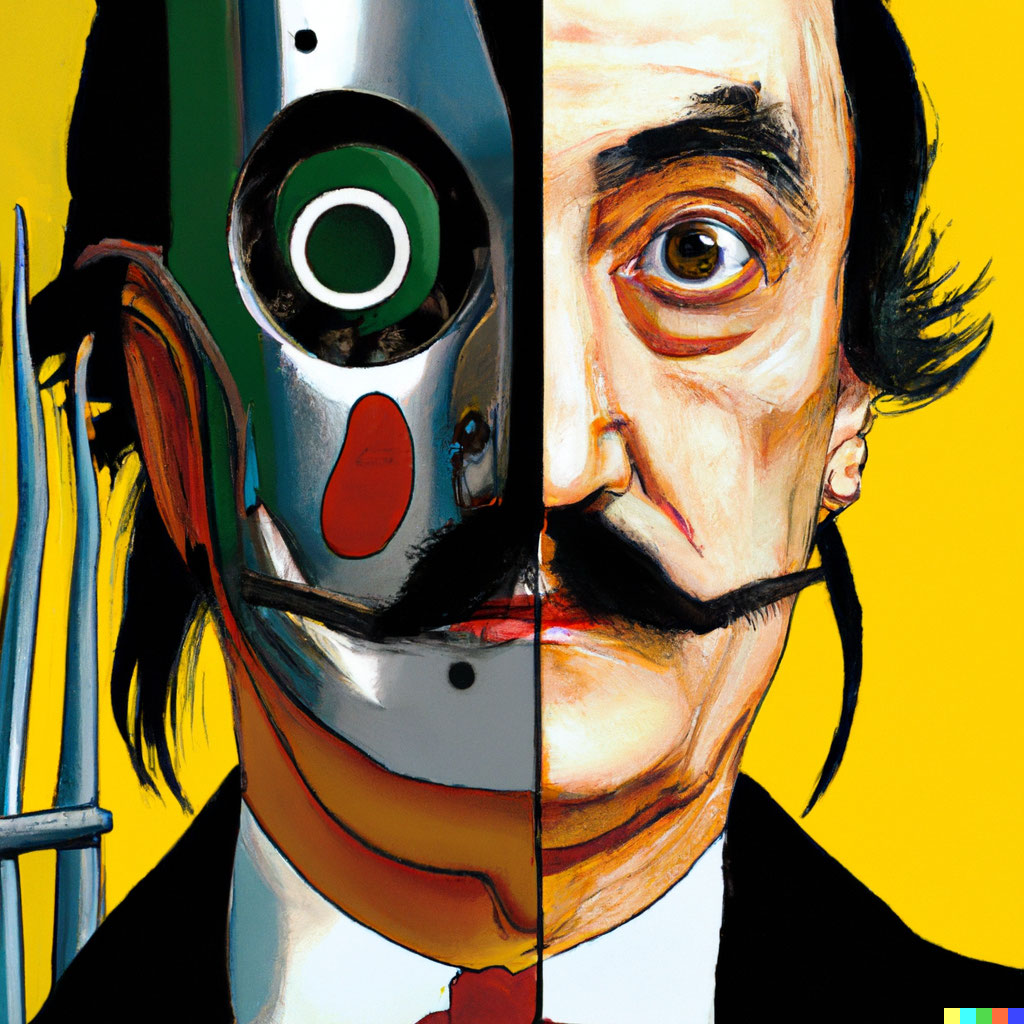} &
    \includegraphics[width=0.31\textwidth]{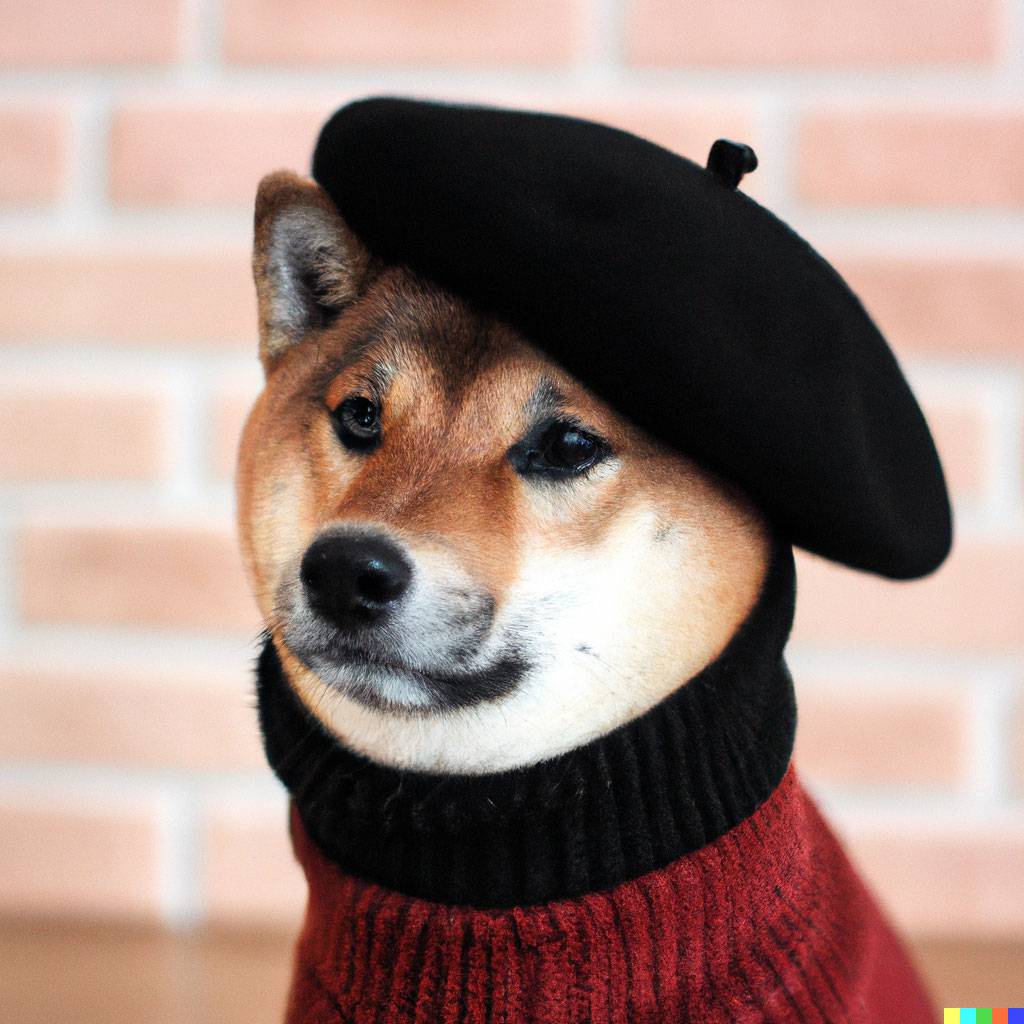} &
    \includegraphics[width=0.31\textwidth]{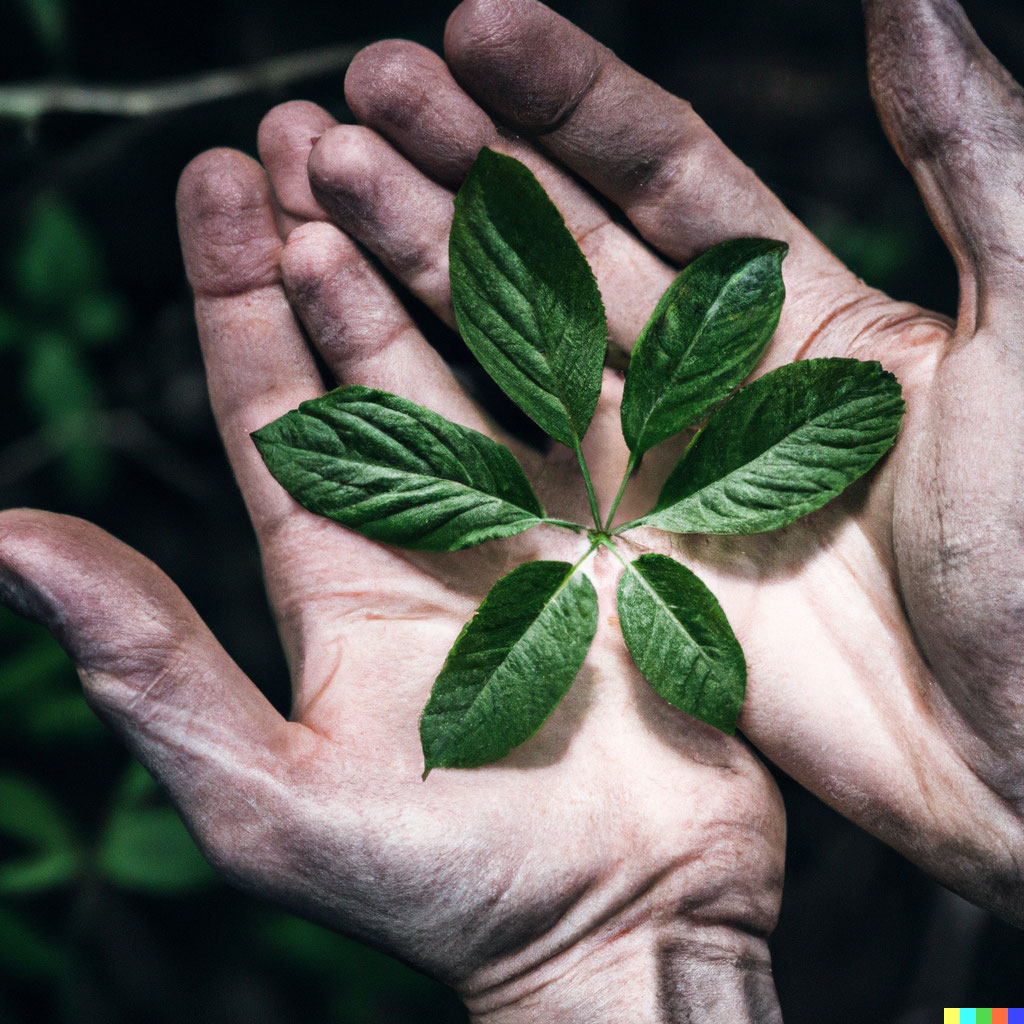} \\
    \scriptsize \makecell{vibrant portrait painting of\\ Salvador Dalí with a robotic \\ half face} &
    \scriptsize \makecell{a shiba inu wearing a beret \\ and black turtleneck} &
    \scriptsize \makecell{a close up of a handpalm with \\
    leaves growing from it} \\
    % \rule{0pt}{0.0pt} \\
    \includegraphics[width=0.31\textwidth]{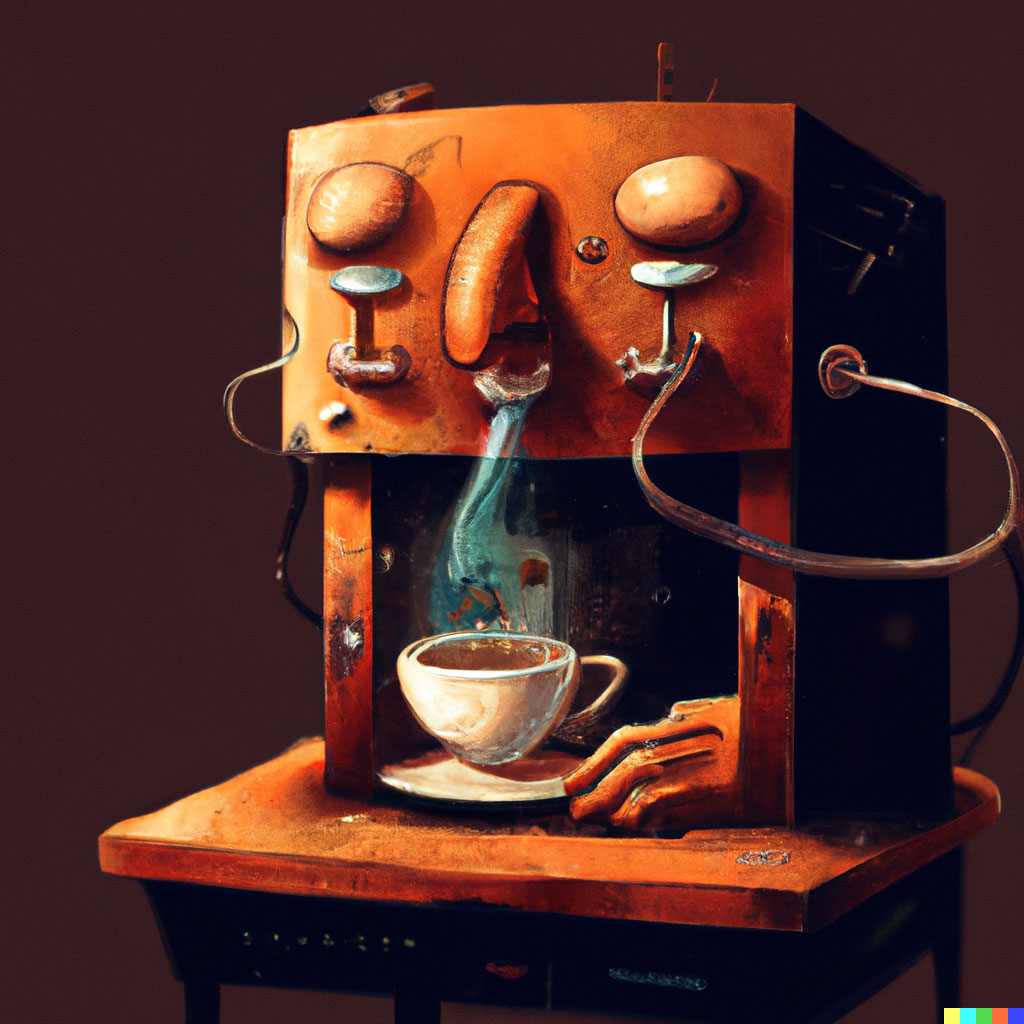} &
    \includegraphics[width=0.31\textwidth]{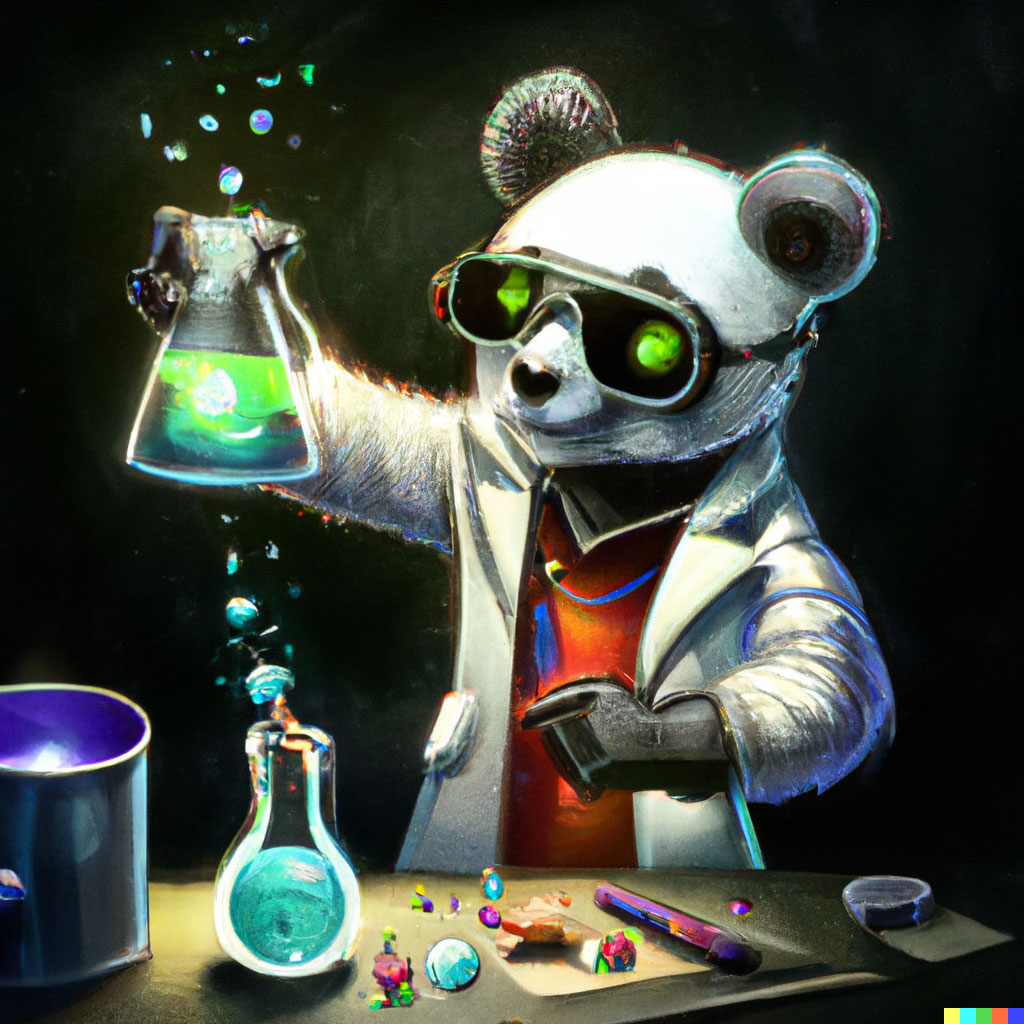} &
    \includegraphics[width=0.31\textwidth]{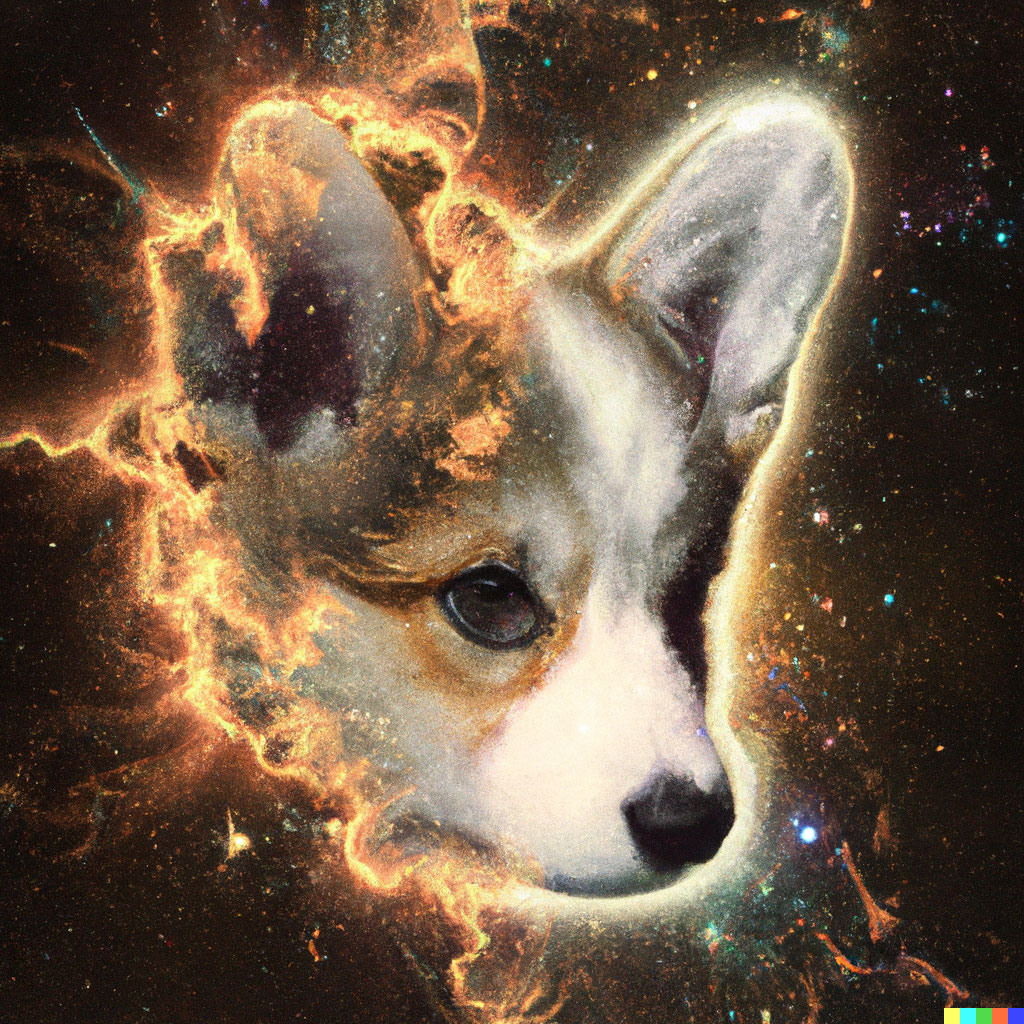} \\
    \scriptsize \makecell{an espresso machine that makes \\
    coffee from human souls, artstation} &
    \scriptsize \makecell{panda mad scientist mixing \\ sparkling 
    chemicals, artstation} &
    \scriptsize \makecell{a corgi's head depicted as \\
    an explosion of a nebula} \\
    % \rule{0pt}{0.0pt} \\
    \includegraphics[width=0.31\textwidth]{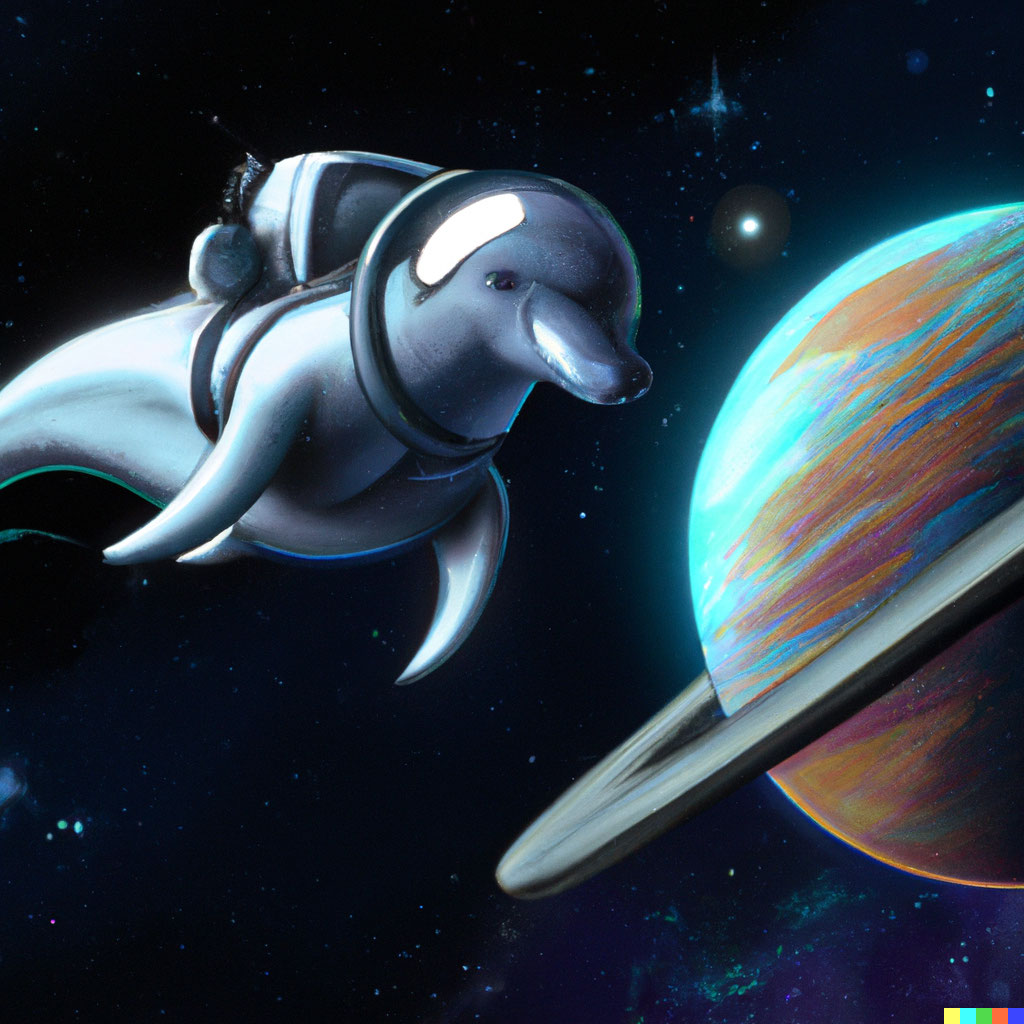} &
    \includegraphics[width=0.31\textwidth]{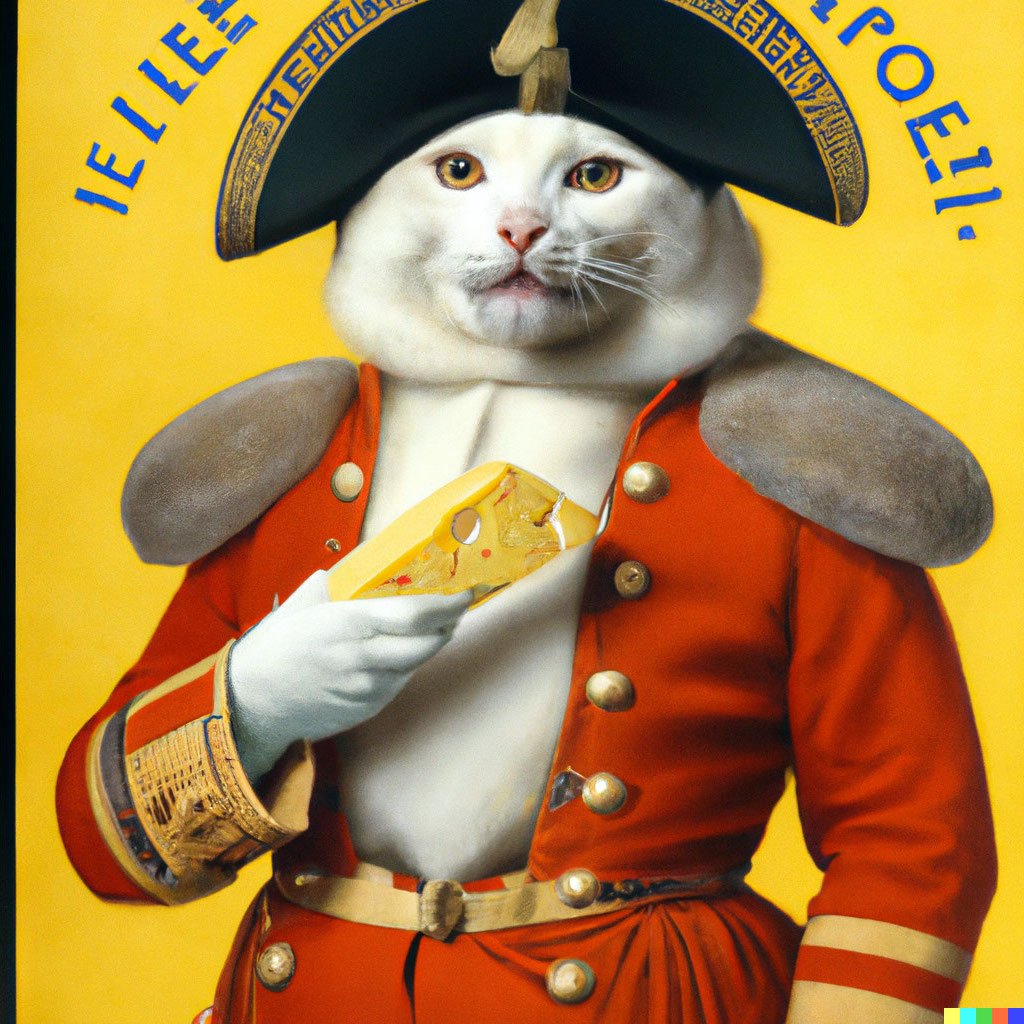} &
    \includegraphics[width=0.31\textwidth]{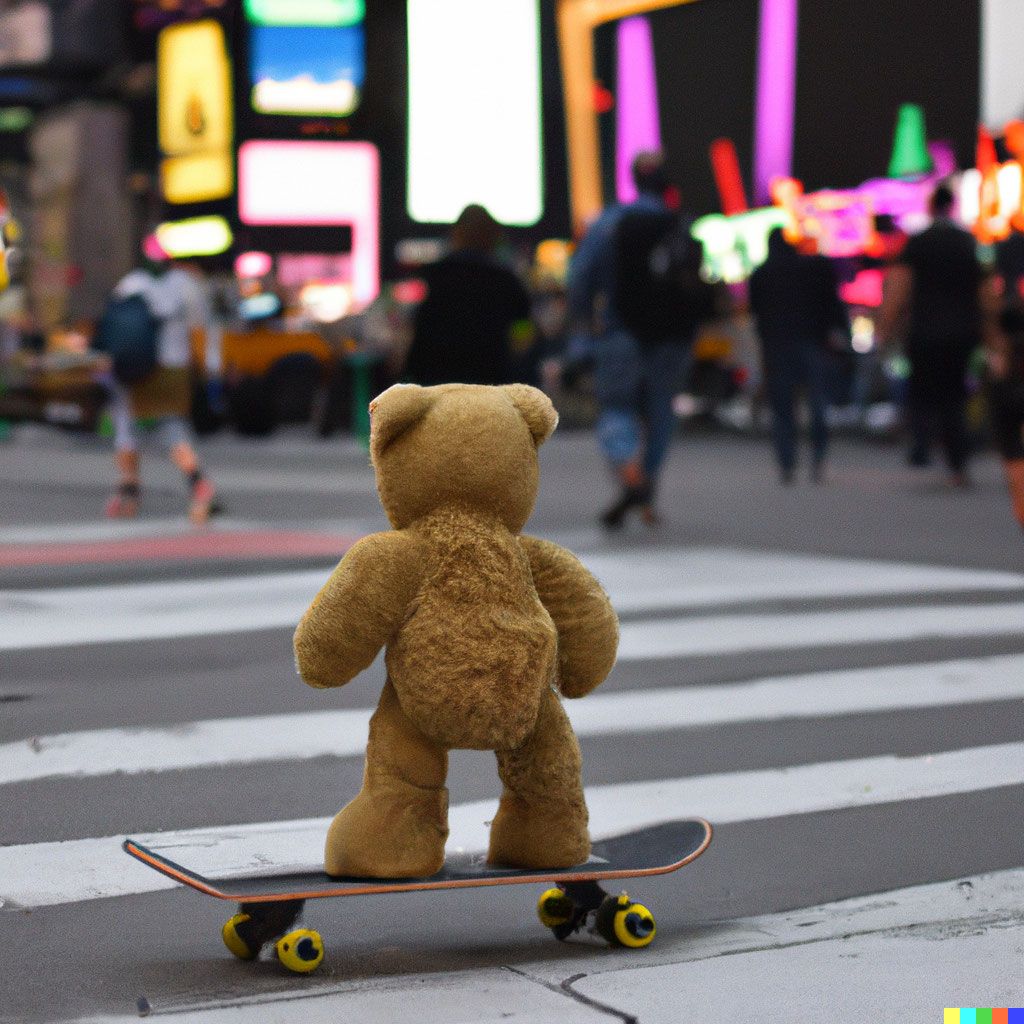} \\
    \scriptsize \makecell{a dolphin in an astronaut \\
    suit on saturn, artstation} &
    \scriptsize \makecell{a propaganda poster depicting a cat \\
    dressed as french emperor\\ napoleon holding a piece of cheese} &
    \scriptsize \makecell{a teddy bear on a skateboard \\
    in times square} \\
    % \rule{0pt}{0.0pt} \\
    \end{tabular}
    \caption{Generated images by text-to-image diffusion models. These images are examples generated by the pioneering model DALL-E2 ~\cite{ramesh2022hierarchical} from OpenAI. Based on user-input text prompts, the model can generate very imaginative images with high fidelity.}
    \label{fig:dalle2_examples}
    % \vskip -0.2in
\end{figure}%

\section{Introduction}\label{sec:introduction}

A picture is worth a thousand words.  Images often convey stories more effectively than text alone. The ability to visualize from text enhances human understanding and enjoyment. Therefore, creating a system that generates realistic images from text descriptions, i.e., the text-to-image (T2I) task,  is a significant step towards achieving human-like or general artificial intelligence.  With the development of deep learning, text-to-image task has become one of the most impressive applications in computer vision~\cite{rombach2022high,ramesh2022hierarchical}.

\begin{figure*}[!tbp]\centering
\includegraphics[width=1.0\linewidth]{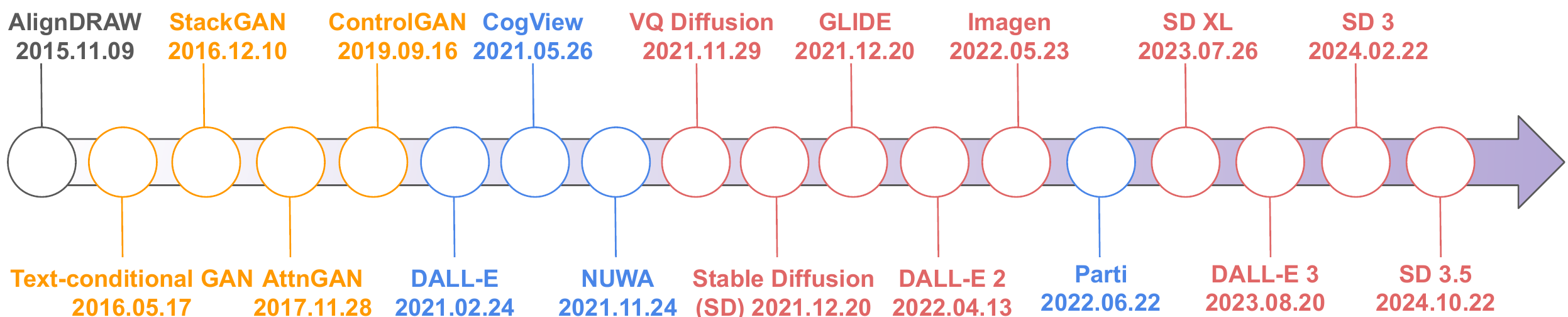}\\
\caption{Representive works on text-to-image task over time. The GAN-based methods, autoregressive methods, and diffusion-based methods are masked in yellow, blue and red, respectively. We abbreviate Stable Diffusion as SD for brevity in this figure. As diffusion-based models have achieved unprecedented success in image generation, this work mainly discusses the pioneering studies for text-to-image generation using diffusion models.
}
\label{fig:timeline}
\end{figure*}

We summarize the timeline of representative studies for text-to-image generation in Figure~\ref{fig:timeline}. AlignDRAW~\cite{mansimov2015generating} marked a significant step by creating images from natural language, albeit with limited realism.  Text-conditional GAN~\cite{reed2016generative} emerged as the first fully end-to-end differential architecture extending from character-level input to pixel-level output, but was always trained on small-scale data. Autoregressive methods further utilize large-scale training data for text-to-image generation,
such as DALL-E~\cite{ramesh2021zero} from OpenAI. However, autoregressive nature makes these methods
~\cite{ramesh2021zero,ding2021cogview,wu2022nuwa,yu2022scaling}
suffer from high computation costs and sequential error accumulation.

More recently, diffusion models (DMs) have emerged as the leading method in text-to-image generation~\cite{nichol2021glide,ramesh2022hierarchical}.
Figure~\ref{fig:dalle2_examples} shows example images generated by the pioneering text-to-image  diffusion model DALL-E2 ~\cite{ramesh2022hierarchical}, demonstrating extraordinary fidelity and imagination.
However, the vast amount of research in this field makes it difficult for readers to learn the key breakthroughs without a comprehensive survey. A branch of existing surveys ~\cite{croitoru2023diffusion,ulhaq2022efficient,cao2022survey} reviews the progress of the diffusion model in all fields, offering a limited introduction specifically on text-to-image synthesis. Other studies~\cite{frolov2021adversarial,ulhaq2022efficient,zhou2021survey} focus on text-to-image tasks using GAN-based approaches, lacking the introduction of diffusion-based methods. 

\begin{figure*}[!tbp]\centering
\includegraphics[width=1.0\linewidth]{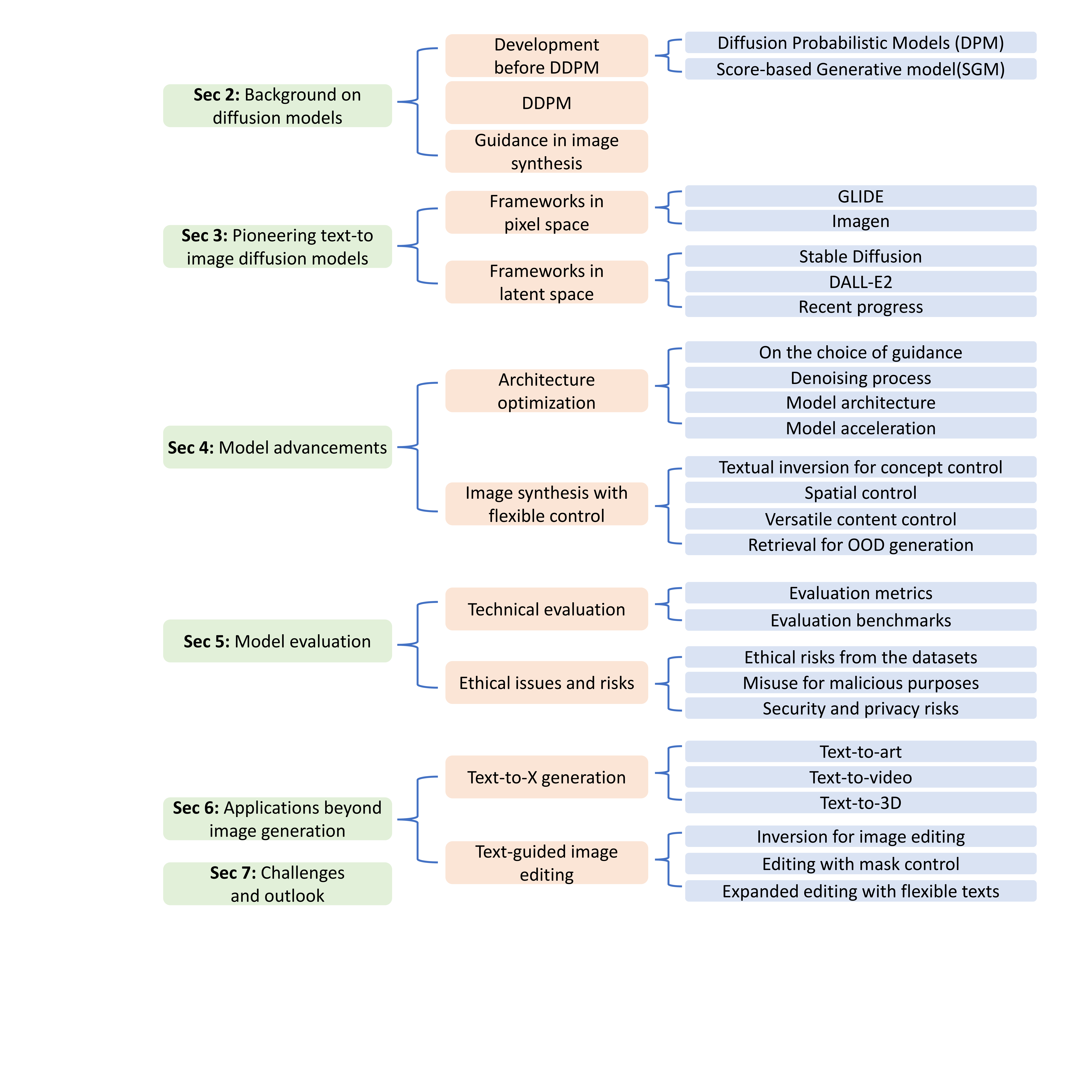}\\
\caption{Paper outline. We summarize each section in this figure. Our work not only offers a comprehensive overview of text-to-image diffusion models, but also provides readers a broader perspective by discussing related areas such as text-to-X generation.
}\label{fig:framework}
\end{figure*}

To our knowledge, this is the first survey to review the progress of diffusion-based text-to-image generation. The rest of the paper is organized as follows. We also summarize the paper outline in Figure~\ref{fig:framework}.
Section~\ref{sec:background} introduces the background of diffusion models. Section~\ref{sec:pioneering_work} covers pioneering studies on text-to-image diffusion models, while Section~\ref{sec:improve_works} discusses the follow-up advancements. Section~\ref{sec:eval_ethical} discusses the evaluation of text-to--image diffusion models from the technical and ethical perspectives. Section~\ref{sec:applications} explores tasks beyond text-to-image generation., such as video generation and 3D object generation. Finally, we discuss challenges and future opportunities in text-to-image generation tasks.

\section{Background on diffusion models}
\label{sec:background}

Diffusion models (DMs), also widely known as diffusion probabilistic models~\cite{sohl2015deep}, are a family of generated models that are Markov chains trained with variational inference~\cite{ho2020denoising}. The learning goal of DM is to reserve a  process of perturbing the data with noise, \textit{i.e.} diffusion, for sample generation~\cite{sohl2015deep,ho2020denoising}. As a milestone work, denoising diffusion probabilistic model (DDPM)~\cite{ho2020denoising} was published in 2020 and sparked an exponentially increasing interest in the community of generative models afterwards.  Here, we provide a self-contained introduction to DDPM by covering the most related progress before DDPM and how unconditional DDPM works with image synthesis as a concrete example. Moreover, we summarize how guidance helps in conditional DM, which is an important foundation for understanding text-conditional DM for text-to-image. 

\subsection{Development before DDPM}
The advent of DDPM~\cite{ho2020denoising} can be mainly attributed to two early attempts: score-based generative models (SGM)~\cite{song2019generative} being investigated in 2019 and diffusion probabilistic models (DPM)~\cite{sohl2015deep} emerging as early as in 2015. Therefore, it is important to revisit the working mechanism of DPM and SGM before we introduce DDPM. 

\textbf{Diffusion Probabilistic Models (DPM).} DPM ~\cite{sohl2015deep} is the first work to model probability distribution by estimating the reversal of Markov diffusion chain which maps data to a simple distribution. Specifically, DPM defines a forward (inference) process  which converts a complex data distribution to a much simpler one, and then learns the mapping by reversing  this diffusion process. Experimental results on multiple datasets show the effectiveness of DPM when estimating complex data distribution. 
DPM  can be viewed as the foundation of DDPM ~\cite{ho2020denoising}, while DDPM optimizes DPM  with improved implementations.

\textbf{Score-based Generative model(SGM).} Techniques for improving score-based generative models have also been investigated in ~\cite{song2019generative}. SGM~\cite{song2019generative} proposes to perturb the data with random Gaussian noise of various magnitudes. With the gradient of log probability density as  score function, SGM generates the samples towards decreasing  noise levels and trains the model by estimating the score functions for noisy data distribution. Despite different motivations, SGM shares a similar optimization objective with DDPM during training, which is also discussed in~\cite{ho2020denoising} that the DDPM under a certain parameterization is equivalent to SGM during training.

\begin{align}
     E_{t \sim \mathcal{U}( 1,T ), \mathbf x_0 \sim q(\mathbf x_0), \epsilon \sim \mathcal{N}(\mathbf{0},\mathbf{I})}{ \lambda(t)  \left\| \epsilon - \epsilon_\theta(\mathbf{x}_t, t) \right\|^2} \label{eq:loss}
\end{align}

\begin{figure*}[!tbp]\centering
\includegraphics[width=0.8\linewidth]{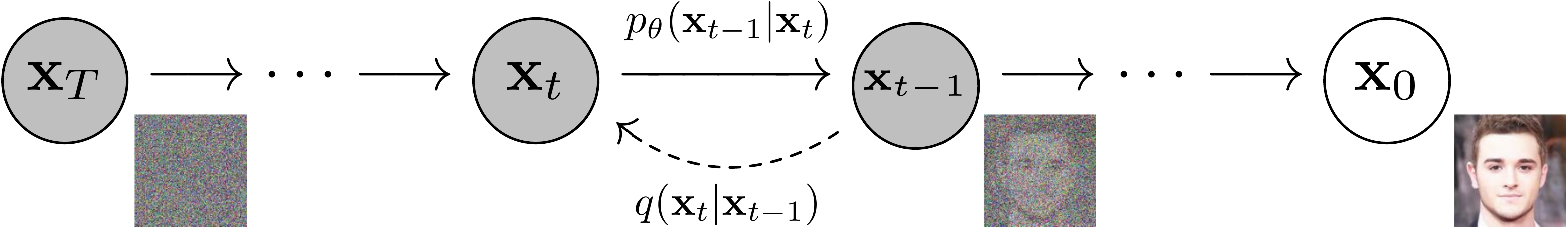}\\
\caption{Diffusion process illustrated in ~\cite{ho2020denoising}. Diffusion models include a forward pass that adds noises to a clean image, and a reverse pass that recovers the clean image from its noisy counterpart.
}
\label{fig:ddpm}
\end{figure*}

\subsection{How does DDPM work for image synthesis?}

Denoising diffusion probabilistic models (DDPMs) are defined as a parameterized Markov chain, which generates images from noise within finite transitions during inference. During training, the transition kernels are learned in a reversed direction of perturbing natural images with noise, where the noise is added to the data in each step and estimated as the optimization target. The diffusion processes are shown in Figure~\ref{fig:ddpm}.

\textbf{Forward pass.} In the forward pass, DDPM is a Markov chain where Gaussian noise is added to data in each step until the images are destroyed. Given a data distribution $\mathbf{x}_0 \sim q(\mathbf{x}_0)$, DDPM generates $\mathbf{x}_T$ successively with  $q(\mathbf{x}_t\mid\mathbf{x}_{t-1})$~\cite{ho2020denoising}:
\begin{align}
q(x_{1:T} | x_0) := \prod_{t=1}^T q( x_t | x_{t-1} ), \label{eq:forwardprocess_1}
\end{align}

\begin{align}
q(x_t|x_{t-1}) := \mathcal{N}(x_t;\sqrt{1-\beta_t} x_{t-1},\beta_t I) \label{eq:forwardprocess_2}
\end{align}

where $T$ and  $\beta_t$ are the diffusion steps and hyper-parameters, respectively. We only discuss the case of Gaussian noise as transition kernels for simplicity, indicated as $\mathcal{N}$ in Eq.~\ref{eq:forwardprocess_2}. With  $\alpha_t := 1 - \beta_t$ and $\bar{\alpha}_t := \prod_{s=0}^{t} \alpha_s$, we can obtain noised image at arbitrary step $t$ as follows~\cite{yang2022diffusionsurvey}:

\begin{align}
q(x_t|x_{0}) := \mathcal{N}(x_t;\sqrt{\bar{\alpha}_t}x_{0},(1 - \bar{\alpha}_t) I) \label{eq:forwardprocess_3}
\end{align}

\textbf{Reverse pass.} With the forward pass defined above, we can train the transition kernels with a reverse process. Starting from $p_\theta(T)$, we hope the generated  $p_\theta(x_0)$ can follow the true data distribution $q(x_0)$. Therefore, the optimization objective of model is as follows(quoted from ~\cite{yang2022diffusionsurvey}):

\begin{align}
     E_{t \sim \mathcal{U}( 1,T ), \mathbf x_0 \sim q(\mathbf x_0), \epsilon \sim \mathcal{N}(\mathbf{0},\mathbf{I})}{ \lambda(t)  \left\| \epsilon - \epsilon_\theta(\mathbf{x}_t, t) \right\|^2} \label{eq:loss}
\end{align}

Considering the optimization objective similarities between DDPM and SGM, thy are unified in~\cite{song2020score} from the perspective of stochastic differential equations, allowing more flexible sampling methods.

\subsection{Guidance in diffusion-based image synthesis}
\label{sec:guidance}

\textbf{Labels improve image synthesis.} Early works on generative adversarial models (GAN) have shown that class labels can improve the quality of generated images by either providing a conditional input or guiding the image synthesis via an auxiliary classifier.  These practices are also introduced to diffusion models:

\textit{Conditional diffusion model:} A conditional diffusion model learns  from additional information  (e.g., class and text) by taking them as model input.

\textit{Guided diffusion model:} During the training of a guided diffusion model, the class-induced gradients (e.g. through an auxiliary classfier) are involved in the sampling process.

\textbf{Classifier-free guidance.} Different from guided diffusion model, ~\cite{ho2022classifier} found that the guidance can be obtained by generative model itself without a classifier, termed as \textit{classifier-free guidance}. Specifically, classifier-free guidance jointly trains a single model with unconditional  score estimator $\epsilon_\theta(x)$ and conditional  $\epsilon_\theta(x, c)$, where $c$ denotes the class label. A null token $\varnothing$ is placed as the class label in the unconditional part, i.e.,  $\epsilon_\theta(x)$ = $\epsilon_\theta(x, \varnothing)$.  Experimental results in~\cite{ho2022classifier} show that classifier-free guidance achieves a trade-off between quality and diversity similar to that achieved by classifier guidance. Without resorting to a classifier, classifier-free diffusion facilitates more modalities, e.g., text in text-to-image, as guidance.

\section{Pioneering text-to-image diffusion models}
\label{sec:pioneering_work}

In this section, we introduce the pioneering text-to-image work based on diffusion models, which can be roughly categorized considering where the diffusion process is conducted, i.e., the pixel space or latent space. The first class of methods generates images directly from the high-dimensional pixel level, including  GLIDE~\cite{nichol2021glide} and Imagen~\cite{saharia2022photorealistic}. Another stream of works propose to first compress the image to a low-dimensional space, and then train the diffusion model on this latent space. Representative methods falling into the class of latent space include Stable Diffusion~\cite{rombach2022high} and DALL-E 2~\cite{ramesh2022hierarchical}. 

\subsection{Frameworks in pixel space}

\begin{figure*}[!tbp]\centering
\includegraphics[width=0.9\linewidth]{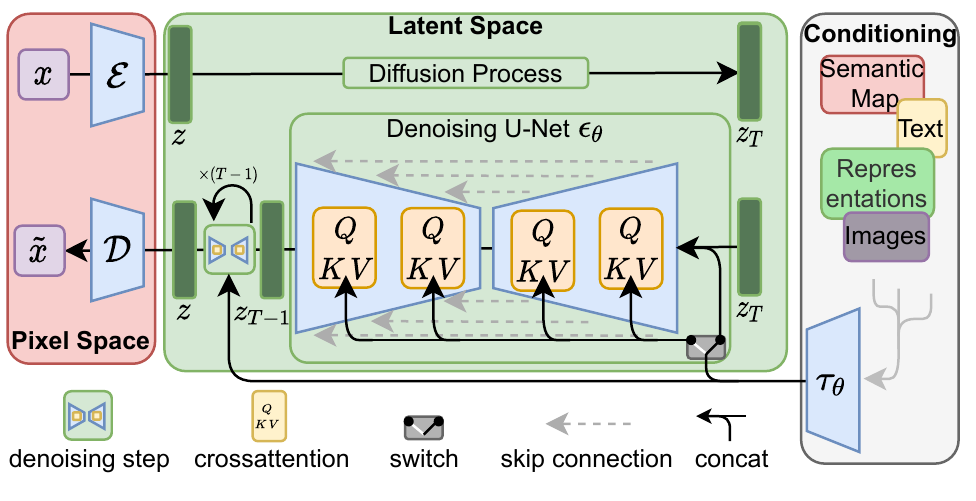}\\
\caption{Model architecture of Stable Diffusion~\cite{rombach2022high}. Stable Diffusion first converts the image to a latent space, where the diffusion process is performed. Stable Diffusion significantly improves the quality and efficiency of image generation compared to prior models.
}
\label{fig:stable_diffusion}
\end{figure*}

\textbf{GLIDE: the first T2I work on DM.}
In essence, text-to-image is text-conditioned image synthesis. Therefore, it is intuitive to replace the class label in class-conditioned DM with \textit{text} for making the sampling generation conditioned on text. As discussed in Sec.~\ref{sec:guidance}, guided diffusion improves the photorealism of samples in conditional DM and its classifier-free variant~\cite{ho2022classifier} facilitates handling free-form prompts. Motivated by this, GLIDE~\cite{nichol2021glide} adopts classifier-free guidance in T2I by replacing original class label with text.  GLIDE~\cite{nichol2021glide} also investigated CLIP guidance but is less preferred by human evaluators than classifier-free guidance for the sample photorealism and caption similarity. As an important component in their framework, the text encoder is set to a transformer with 24 residual blocks with a width of 2048 (roughly 1.2B parameters). Experimental results show that GLIDE~\cite{nichol2021glide} outperforms DALL-E~\cite{ramesh2021zero}  in both FID  and human evaluation.

\textbf{Imagen: encoding text with pretrained  language model.} Following GLIDE~\cite{nichol2021glide}, Imagen~\cite{saharia2022photorealistic} adopts classifier-free guidance for image generation.  A core difference between GLIDE and Imagen lies in their choice of text encoder.  Specifically, GLIDE trains the text encoder together with the diffusion prior with paired image-text data, while Imagen~\cite{saharia2022photorealistic} adopts a pretrained and frozen large language model as the  text encoder.  Since the text-only corpus is significantly larger than paired image-text data, such as 800GB used in T5~\cite{raffel2020exploring}, the pretrained large language models are exposed to text with a rich and wide distribution. With different T5~\cite{raffel2020exploring} variants  as the text encoder, ~\cite{saharia2022photorealistic} reveals that increasing the size of language model improves  the image fidelity and image-text alignment more than enlarging the diffusion model size in Imagen.
Moreover,  freezing the weights of pretrained encoder facilitates offline text embedding, which reduces negligible computation burden to the online training of the text-to-image diffusion prior.  

\subsection{Frameworks in latent space}

\begin{figure*}[!tbp]\centering
\includegraphics[width=0.9\linewidth]{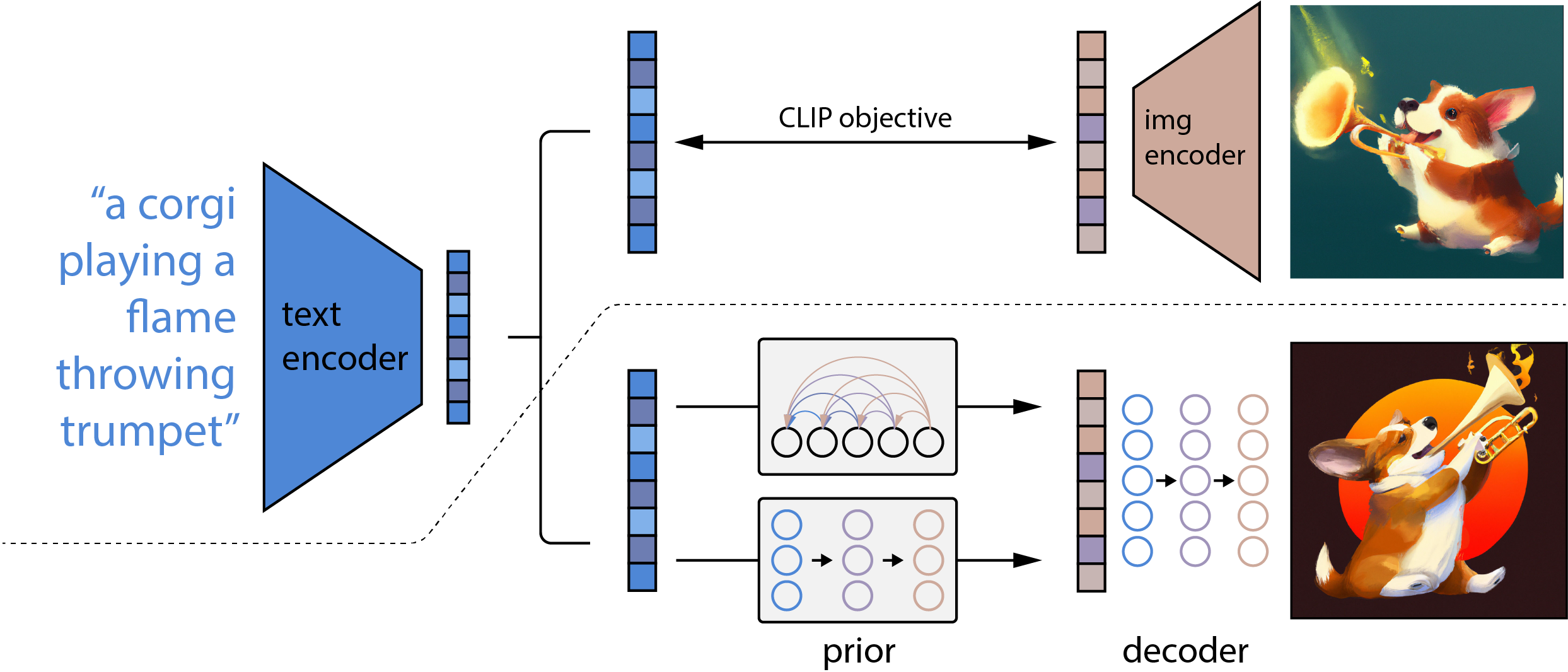}\\
\caption{Model architecture of DALLE-2~\cite{ramesh2022hierarchical}. DALLE-2 uses CLIP~\cite{radford2021learning} model to project the image and text to latent space.
}
\label{fig:dit}
\end{figure*}

\textbf{Stable diffusion: a milestone work on latent space.} A representative framework that trains the diffusion models on latent space is Stable Diffusion, which is a scaled-up version of Latent Diffusion Model (LDM)~\cite{rombach2022high}. Following Dall-E~\cite{ramesh2021zero} that adopts a  VQ-VAE to learn a visual codebook, Stable diffusion applies VQ-GAN  for the latent representation in the first stage. Notebly, VQ-GAN improves VQ-VAE by adding an adversarial objective to increase the naturalness of synthesized images. With the pretrained VAE, stable diffusion reverses a forward diffusion process that perturbs latent space with noise. Stable diffusion also introduces cross-attention as general-purpose conditioning for various condition signals like text. The model architecture of Stable Diffusion is shown in Figure~\ref{fig:stable_diffusion}. Experimental results in ~\cite{rombach2022high} highlight that performing diffusion modeling on the latent space significantly outperforms that on the pixel space in terms of complexity reduction and detail preservation. A similar approach has also been investigated in VQ-diffusion with a mask-then-replace diffusion strategy. Resembling the finding in pixel-space method, classifier-free guidance also significantly improves the text-to-image diffusion models in latent space~\cite{rombach2022high}.

\textbf{DALL-E2: with multimodal latent space.} Another stream of text-to-image diffusion models in latent space relies on multimodal contrasitve models~\cite{radford2021learning}, where image embedding and text encoding are matched in the same representation space. For example, CLIP~\cite{radford2021learning} is a pioneering work learning the multimodal representations and has been widely used in numerous text-to-image models~\cite{ramesh2022hierarchical}. A representative work applying CLIP is DALL-E 2, also known as unCLIP~\cite{ramesh2022hierarchical}, which adopts the CLIP text encoder but inverts the CLIP image encoder with a diffusion model that generates images from CLIP latent space. 
 Such a combination of  encoder and decoder resembles the structure of VAE adopted in LDM, even though the inverting decoder is non-deterministic~\cite{ramesh2022hierarchical}. 
Therefore, the remaining task is to train a prior to bridge the gap between CLIP text and image latent space, and we term it as \textit{text-image latent prior} for brevity. DALL-E2 ~\cite{ramesh2022hierarchical} finds that this prior can be learned by either autoregressive method or diffusion model, but diffusion prior achieves superior performance. Moreover, experimental results show that removing this \textit{text-image latent prior} leads to a performance drop by a large margin~\cite{ramesh2022hierarchical}, which highlights the importance of learning the \textit{text-image latent prior}. We show image examples generated by DALLE-2 in Figure~\ref{fig:dalle2_examples}.

\textbf{Recent progress of Stable Diffusion and DALL-E family.}  Since the publication of Stable Diffusion~\cite{rombach2022high}, multiple versions of models have been released, including Stable Diffusion 1.4, 1.5, 2.0, 2.1, XL, and 3. Starting from  Stable Diffusion 2.0~\cite{stable_diffusion_2.0}, a notable feature is negative prompts, which allow users to specify what they do not wish to generate in the output image. Stable Diffusion XL~\cite{stable_diffusion_xl} enhances capabilities beyond previous versions by incorporating a larger Unet architecture, leading to improved abilities such as face generation, richer visuals, and more impressive aesthetics. Stable Diffusion 3 is built on diffusion transformer architecture~\cite{peebles2023scalable} and use two separate sets of weights to model text and image modality. Stable Diffusion 3 improve overall comprehension and typography of generated images.  On the other hand, the evolution of the DALL-E model has progressed from the autoregressive DALL-E~\cite{ramesh2021zero}, to the diffusion-based DALL-E2 ~\cite{ramesh2022hierarchical}, and most recently, DALLE-3~\cite{openai-dall-e-3}. Integrated into the GPT-4 API, DALLE-3 showcases superior performance in capturing intricate nuances and details.

\section{Model advancements}
\label{sec:improve_works}
Numerous works attempt to improve the text-to-image diffusion models, which we roughly categorize into architecture optimization and versatile use.
\subsection{Architecture optimization}

\textbf{On the choice of guidance.} Beyond the classifier-free guidance, some works~\cite{nichol2021glide} have also explored cross-modal guidance with CLIP~\cite{radford2021learning}. Specifically, GLIDE~\cite{nichol2021glide} finds that CLIP-guidance underperforms the classifier-free variant of guidance. By  contrast, another work UPainting~\cite{li2022upainting}
points out that lacking of a large-scale transformer language model makes these models with CLIP guidance difficult to encode text prompts and generate complex scenes with details. By combing large language model  and cross-modal matching models, UPainting~\cite{li2022upainting} significantly  improves the sample fidelity and image-text alignment of generated images. The general image synthesis capability enables  UPainting~\cite{li2022upainting} to generate images in both simple and complex scenes.

\textbf{Denoising process.} By default, DM during inference repeats the denoising process on the same denoiser model, which makes sense for an unconditional image synthesis since the goal is only to get a high-fidelity image. In the task of text-to-image synthesis, the generated image is also required to align with the text, which implies that the denoiser model has to make a trade-off between these two goals. Specifically, two recent works~\cite{feng2023ernie,balaji2022ediffi} point out a phenomenon: the early sampling stage strongly relies on the text prompt for the goal of aligning with the caption, but the later stage focuses on improving image quality while almost ignoring the text guidance. Therefore, they abort the practice of sharing model parameters during the denoising process and propose to adopt multiple denoiser models which are specialized for different generation stages. Specifically, ERNIE-ViLG 2.0~\cite{feng2023ernie} also mitigates the problem of object-attribute by the guidance of a text parser and object detector, improving the fine-grained semantic control.

\begin{figure*}[!tbp]\centering
\includegraphics[width=1.0\linewidth]{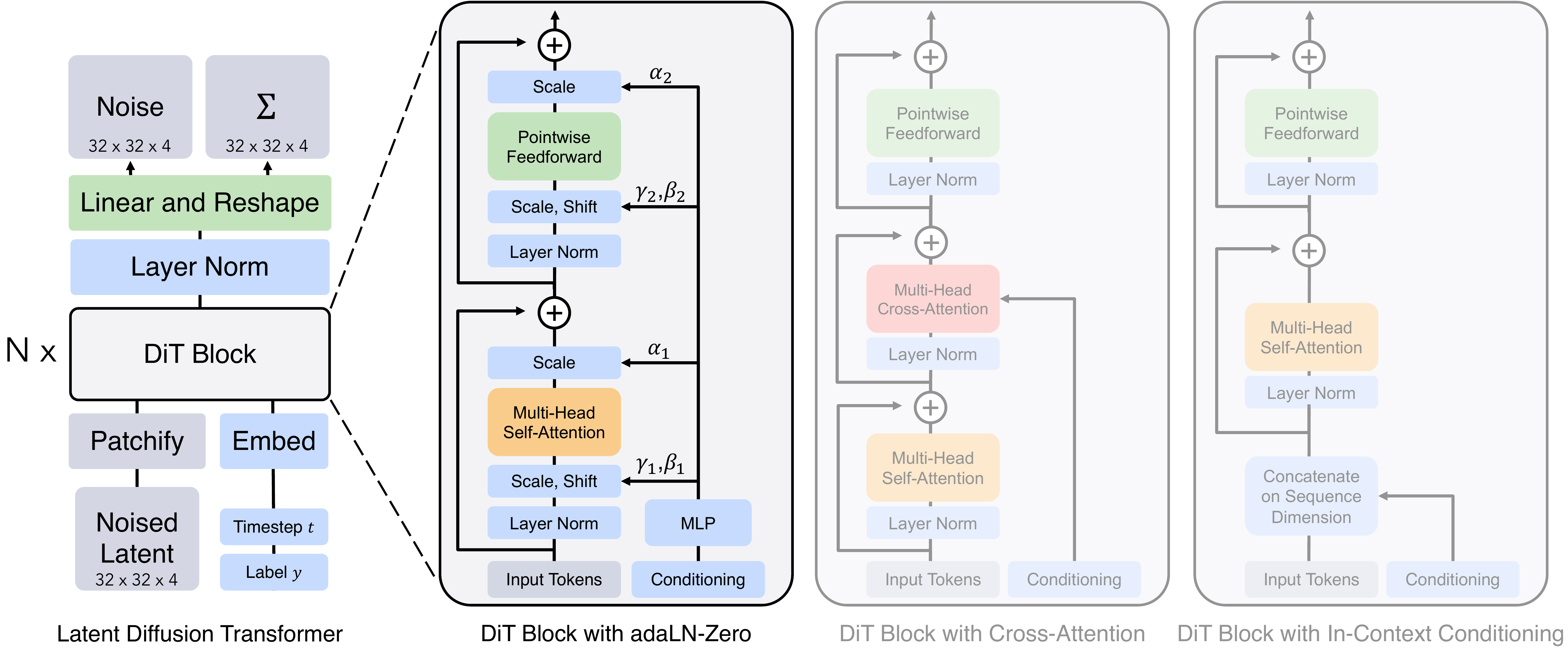}\\
\caption{Transformer architecture of DiT~\cite{peebles2023scalable}.  DiT  trains conditional latent diffusion models with transformer blocks. Adaptive layer norm works best among all block types.
}
\label{fig:dit}
\end{figure*}

\textbf{Model architecture.} A branch of studies enhances  text-to-image generation by improving the denoising model. For instance, Free-U~\cite{si2024freeu}  strategically re-weights the contributions sourced from the U-Net’s skip connections and backbone feature maps, which improves image generation quality without additional training or fine-tuning. The pioneering work DiT~\cite{peebles2023scalable} proposes a diffusion transformer architecture as the denoising model of diffusion models, which replaces the commonly-used U-Net backbone (see Figure~\ref{fig:dit}).  Pixart-$\alpha$~\cite{chen2023pixart} is a pioneering work that adopts a transformer-based backbone and supports high-resolution image synthesis up to 1024 × 1024 resolution with low training cost.

\textbf{Model acceleration.} Diffusion models have achieved great success in image generation, outperforming GAN. However, one drawback of diffusion models is their slow sampling process, which requires hundreds or thousands of iterations to generate an image. 
 V-prediction~\cite{salimans2021progressive} improves the sampling speed by distilling a pre-trained diffusion model with N-step DDIM sampler to a new model of N/2 sampling steps, without hurting generation quality.  Flow Matching (FM)~\cite{lipmanflow} finds that employing FM with diffusion paths results in a more robust and stable alternative for training diffusion models. A recent work REPresentation Alignment (REPA) ~\cite{yu2024representation}  emphasizes the key role of representations in 
training large-scale diffusion models, and introduces the representations from self-supervised models (DINO v2~\cite{oquab2024dinov2}) to the training of diffusion models like DiT~\cite{peebles2023scalable} and SiT~\cite{ma2024sit}. REPA~\cite{yu2024representation} achieves significant acceleration results by speeding up SiT~\cite{ma2024sit}
training by over 17.5×.

\subsection{Image synthesis with flexible control}

\begin{figure*}[!tbp]\centering
\includegraphics[width=1.0\linewidth]{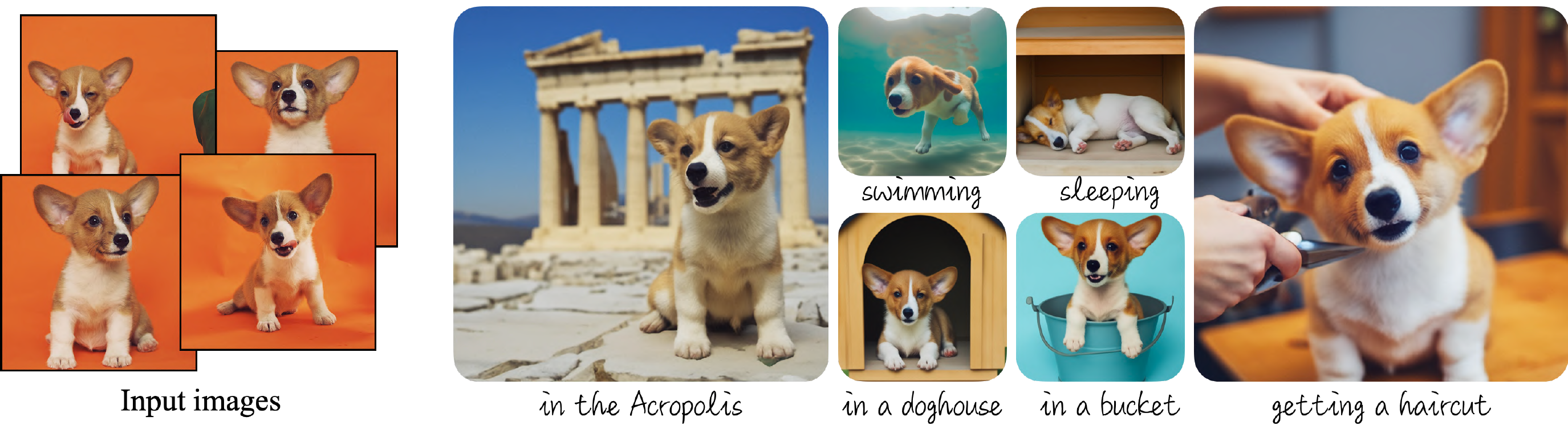}\\
\caption{Textual inversion for concept control in Dreambooth~\cite{ruiz2022dreambooth}. Based on the user input images, Dreambooth~\cite{ruiz2022dreambooth} finetunes a pretrained model to learn the key concept of subject in input images. The users can further control the status of the subject by prompts such as  ``getting a haircut". }
\label{fig:dreambooth}
\end{figure*}

\textbf{Textual inversion for concept control.} Pioneering works on text-to-image generation ~\cite{nichol2021glide,saharia2022photorealistic,rombach2022high,ramesh2022hierarchical} rely on natural language  to describe the content and styles of generated images. However, there are cases when the text cannot exactly describe the desired semantics by users, e.g., generating a new subject.  In order to synthesize novel scenes with certain concepts or subjects, ~\cite{gal2022image,ruiz2022dreambooth}  introduces several reference images with the desired concepts, then inverts the reference images to the textual descriptions. Specifically, ~\cite{gal2022image} inverts the shared concept in a couple of reference images into the text (embedding) space, i.e. "pseudo-words". The generated "pseudo-words" can be used for personalized generation. DreamBooth~\cite{ruiz2022dreambooth} adopts a similar technique and mainly differs by fine-tuning (instead of freezing) the pretrained DM model for preserving key visual features from the subject identity. With the learned subject, the  DreamBooth~\cite{ruiz2022dreambooth} allows users to control the status of subject in the generated images by specifying in the input prompt. Figure~\ref{fig:dreambooth} shows generated images by DreamBooth with dog images as well as the prompt as model inputs.
Textual inversion has also been applied in other applications, such as to control the spatial relationship in ~\cite{huang2023reversion}.

\begin{figure*}[!tbp]\centering
\includegraphics[width=1.0\linewidth]{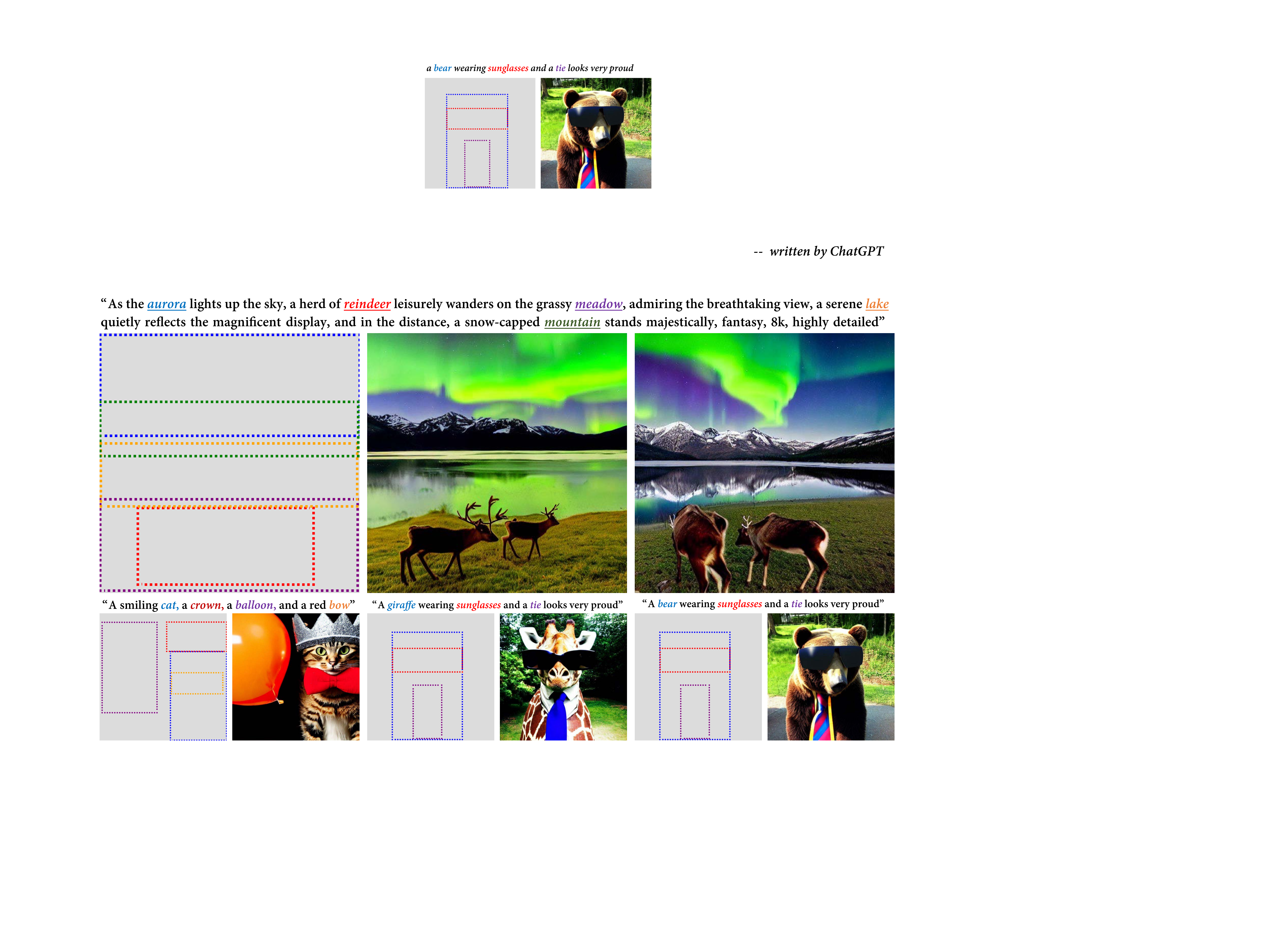}\\
\caption{Spatial control in  BoxDiff~\cite{zhang2023adding}. BoxDiff enables to control the layout of generated images with provided boxes or scribbles.}

\label{fig:boxdiff}
\end{figure*}

\textbf{Spatial control.} Despite their unprecedented high image fidelity and caption similarity, most text-to-image DMs like Imagen~\cite{saharia2022photorealistic} and DALL-E2~\cite{ramesh2022hierarchical} do not provide fine-grained control of spatial layout. To this end, SpaText~\cite{avrahami2023spatext} introduces spatio-textual (ST) representation which can be included to finetune a SOTA DM by adapting its decoder. Specifically, the new encoder conditions both local ST and existing global text. Therefore, the core of SpaText~\cite{avrahami2023spatext} lies in ST where the diffusion prior in trained separately to convert the image embeddings in CLIP to its text embeddings. During training, the ST is generated directly by using the CLIP image encoder taking the segmented image object as input. A concurrent work~\cite{voynov2022sketch} proposes to realize fine-grained local control through a simple sketch image. Core to their approach is a Latent Guidance Predictor (LGP)that is a pixel-wise MLP mapping the latent feature of a noisy image to that of its corresponding sketch input. After being trained (see~\cite{voynov2022sketch} for more training details), the LGP can be deployed to the pretrained text-to-image DM without the need for fine-tuning. Other representative studies for spatial control includes  BoxDiff~\cite{xie2023boxdiff} that uses the provided box or scribble to control the layout of generated images, as shown in Figure~\ref{fig:boxdiff}.

\begin{figure*}[!tbp]\centering
\includegraphics[width=1.0\linewidth]{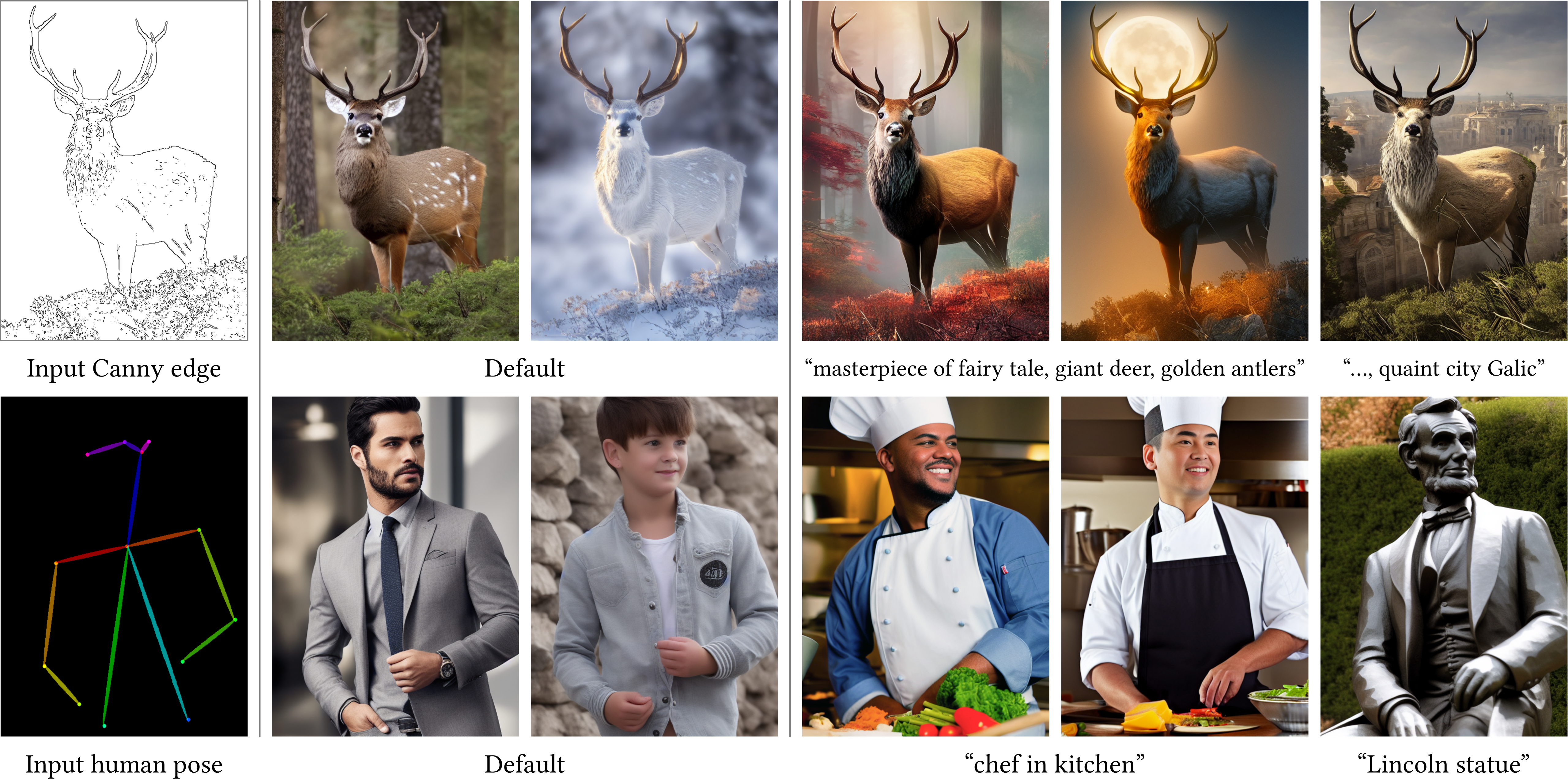}\\
\caption{Control Stable Diffusion with conditions~\cite{zhang2023adding}. ControlNet~\cite{zhang2023adding} allows users to specify conditions, such as canny edges,  in image generation of large-scale pretrained diffusion models. For example, the default prompt is ``a high-quality, detailed, and professional image'', while the users can add conditions such as ``quaint city Galic''.}

\label{fig:controlnet}
\end{figure*}
\textbf{Versatile content control.} ControlNet ~\cite{zhang2023adding} has achieved great attention due to its powerful ability to add various conditioning controls to large pretrained models. ControlNet ~\cite{zhang2023adding} reuses the pretrained encoding layers as a strong backbone, and a zero convolution architecture is proposed to ensure no harm noise could affect the finetuning. ControlNet ~\cite{zhang2023adding} achieves outstanding results with various conditioning signals, such as edges, depth, and segmentation. Figure~\ref{fig:controlnet} shows an example from ~\cite{zhang2023adding} that use canny edge and human as condition to control the image generation of Stable Diffusion model. There are also other widely used methods unifies various signals in one model for content control, such as T2i-adapter~\cite{mou2024t2i}, Uni-ControlNet~\cite{zhao2024uni}, GLIGEN~\cite{li2023gligen}, and Composer~\cite{huang2023composer}.  HumanSD~\cite{ju2023humansd} and HyperHuman~\cite{liu2024hyperhuman} focuses on the generation of human images by taking human skeleton as model inputs.

\textbf{Retrieval for out-of-distribution generation.} State-of-the-art text-to-image models assume sufficient exposure to descriptions of common entities and styles from training. This assumption breaks down with rare entities or vastly different styles, leading to performance drops. To counter this, several studies~\cite{blattmann2022retrieval, sheynin2022knn, rombach2022text, chen2022re} use external databases for retrieval, a semi-parametric approach adapted from NLP~\cite{khandelwal2019generalization, guu2020retrieval} and GAN-based synthesis~\cite{li2022memory}. Retrieval-Augmented Diffusion Models (RDMs)~\cite{blattmann2022retrieval} use $k$-nearest neighbors (KNN) based on CLIP distance for enhanced diffusion guidance, while KNN-diffusion~\cite{sheynin2022knn} improves quality by adding text embeddings. Re-Imagen~\cite{chen2022re} refines this with a single-stage framework, retrieving both images and text in latent space, outperforming KNN-diffusion on the COCO benchmark.

\section{Model evaluation}
\label{sec:eval_ethical}

\subsection{Technical evaluation}
\label{sec:tech_eval}

\textbf{Evaluation metrics.} A common metric to evaluate image quality quantitatively is  Fréchet Inception Distance (FID), which measures the Fréchet distance (also known as Wasserstein-2 distance) between synthetic and real-world images.  We summarize the evaluation results of representative methods on  MS-COCO dataset in Table ~\ref{tab:results_fid} for reference. The smaller the FID is, the higher the image fidelity. To measure the text-image alignment, CLIP scores are widely applied, which trades off against FID. There are also other metrics for text-to-image evaluation, including Inception score (IS)~\cite{salimans2016improved} for image quality and R-precision for text-to-image generation.

\begin{wraptable}[12]{ht}{0.5\textwidth}
\vspace{-22pt}
\caption{Image quality comparison of autoregressive and diffusion models. Diffusion models outperforms autoregressive models in image quality, with lower FID on MS-COCO dataset.}
\label{tab:results_fid}
\resizebox{1.0\linewidth}{!}{
\begin{tabular}{l|l|llllll}
\hline
& model     &FID  ($\downarrow$)\\
\hline 
Autoregressive & CogView~\cite{ding2021cogview}& 27.10  \\
& LAFITE ~\cite{zhou2022towards}  & 26.94 \\
& DALLE~\cite{ramesh2021zero} & 17.89 \\
\hline 
Diffusion models &  GLIDE~\cite{nichol2021glide}  & 12.24   \\
& Imagen~\cite{saharia2022photorealistic}   & 7.27  \\
 & Stable Diffusion~\cite{rombach2022high}  & 12.63  \\
% &  VQ-Diffussion~\cite{gu2022vector}  & 13.86 \\
&  DALL-E 2~\cite{ramesh2022hierarchical}  &  10.39  \\
& Upainting~\cite{li2022upainting} & 8.34  \\
& ERNIE-ViLG 2.0~\cite{feng2023ernie} & 6.75 \\
& eDiff-I ~\cite{balaji2022ediffi} & 6.95 \\
\hline 
\end{tabular}}
\end{wraptable}

\textbf{Evaluation benchmarks.} Apart from the automatic metrics discussed above, multiple works involve human evaluation and propose their new evaluation benchmarks ~\cite{cho2022dall,saharia2022photorealistic,yu2022scaling,li2022upainting,petsiuk2022human,chen2022re,liao2022artbench}. We summarize representative benchmarks in Table~\ref{tab:benchmark}. For a better evaluation of fidelity and text-image alignment, DrawBench\cite{saharia2022photorealistic}, PartiPropts~\cite{yu2022scaling}  and UniBench~\cite{li2022upainting} ask the human raters to compare generated images from different models. Specifically,  UniBench~\cite{li2022upainting} proposes to evaluate the model on both simple and complex scenes and includes both Chinese and English prompts.  PartiPropts~\cite{yu2022scaling} introduces a diverse set of over 1600 (English) prompts and also proposes a challenge dimension that highlights why this prompt is difficult. To evaluate the model from various aspects,   PaintSKills~\cite{cho2022dall}  evaluates the \textit{visual reasoning skills} and \textit{social biases} apart from image quality and text-image alignment. However, PaintSKills~\cite{cho2022dall} only focuses on unseen object-color and object-shape scenario~\cite{li2022upainting}.  EntityDrawBench~\cite{chen2022re} further evaluates the model with various infrequent entities in different scenes. Compared to PartiPropts~\cite{yu2022scaling} with prompts at different difficulty levels, Multi-Task Benchmark~\cite{petsiuk2022human} proposes thirty-two tasks that evaluate different capabilities and divides each task into three difficulty levels.

\subsection{Ethical issues and risks}

\textbf{Ethical risks from the datasets.} Text-to-image generation is a highly data-driven task, and thus models trained on large-scale unfiltered data may suffer from even reinforce the biases from the dataset, leading to ethical risks. ~\cite{schramowski2022safe} finds a large amount of inappropriate content in the generated images by Stable diffusion~\cite{rombach2022high} (e.g., offensive, insulting, or threatening information), and first establishes a new test bed to evaluate them. Moreover, it proposes Safe Latent Diffusion, which successfully removes and suppresses inappropriate content with additional guidance.
Another ethical issue,  the fairness of social group, is studied in ~\cite{struppek2022biased, bansal2022well}. Specifically, ~\cite{struppek2022biased} finds that simple homoglyph replacements in the text descriptions can induce culture bias in models, i.e., generating images from different cultures. ~\cite{bansal2022well} introduce an Ethical NaTural Language Interventions
in Text-to-Image GENeration (ENTIGEN)
benchmark dataset, which can evaluate the change of generated images with ethical interventions by three axes: gender, skin color, and culture. With intervented text prompts, ~\cite{bansal2022well} improves diffusion models (e.g., Stable diffusion~\cite{rombach2022high}) from the social diversity perspective. Fair Diffusion~\cite{friedrich2023fair} evaluates the fairness problem of diffusion models and mitigates this problem at the deployment stage of diffusion models. Specifically, Fair Diffusion~\cite{friedrich2023fair} instructs the diffusion models on fairness with textual guidance. Another work~\cite{bianchi2023easily} finds that a broad range of prompts to text-to-image diffusion models could produce stereotypes, such as simply mentioning traits, descriptors, occupations, or objects.

\begin{table*}[!tbp] \centering 
\caption{Benchmarks for text-to-image generation task.}
\label{tab:benchmark}
\resizebox{1.0\linewidth}{!}{
\begin{tabular}{lllllllllllllll}
\hline
Benchmark   & Measurement & Metric  & Auto-eval &  Human-eval     &  Language\\
\hline 
DrawBench\cite{saharia2022photorealistic}  &  Fidelity, alignment& User preference rates &  N  &  Y   & English \\
UniBench~\cite{li2022upainting}  & Fidelity, alignment  & User preference  rates & N &  Y    &  English, Chinese\\
PartiPrompts~\cite{yu2022scaling} & Fidelity, alignment & Qualitative  & N &  Y   & English\\
PaintSKills~\cite{cho2022dall} & Visual reasoning skills, social biases & Statistics  & Y  &  Y   &  English\\
EntityDrawBench~\cite{chen2022re} & Entity-centric  faithfulness & Human rating  & N &  Y   &  English\\
Multi-Task Benchmark~\cite{petsiuk2022human} & Various capabilities & Human rating  & N &  Y   &    English\\
\hline 
\end{tabular}}
\end{table*}

\textbf{Misuse for malicious purposes.} Text-to-image diffusion models have shown their power in generating high-quality images. However, this also raises great concern that the generated images may be used for malicious purposes, e.g., falsifying electronic evidence~\cite{sha2023fake}. DE-FAKE~\cite{sha2023fake} is the first to conduct a systematical study on visual forgeries of the text-to-image diffusion models, which aims to distinguish generated images from the real ones, and also further track the source model of each fake image.
To achieve these two goals, DE-FAKE~\cite{sha2023fake} analyzes from visual modality perspective,  and finds that images generated by different diffusion models share common features and also present unique model-wise fingerprints. 
Two concurrent works \cite{ricker2022towards,corvi2023detection} approach the detection of faked images both by evaluating the existing detection methods on images generated by the diffusion model, and also analyzing the frequency discrepancy of images by  GAN and diffusion models.
\cite{ricker2022towards,corvi2023detection} find that the performance of existing detection methods drops significantly on generated images by diffusion models compared to GAN. Moreover, \cite{ricker2022towards} attributes the failure of existing methods to the mismatch of high frequencies between images generated by diffusion models and GAN.  Another work ~\cite{ghosh2022can} discusses the concern of artistic image generation from the perspective of artists. Although agreeing that the artistic image generation may be a promising modality for the development of art, ~\cite{ghosh2022can} points out that the artistic image generation may cause plagiarism and profit shifting (profits in the art market shift from artists to model owners) problems if not properly used.

\textbf{Security and privacy risks.} While text-to-image diffusion models have attracted great attention, the security and privacy risks have been neglected so far. Two pioneering works~\cite{struppek2022rickrolling,wu2022membership} discuss the backdoor attack and privacy issues, respectively. Inspired by the findings in 
~\cite{struppek2022biased} that a simple word replacement can invert culture bias to models, Rickrolling the Artist~\cite{struppek2022rickrolling} proposes to inject the backdoors into the pre-trained text encoders, which will force the generated image to follow a specific description or include certain attributes if the trigger exists in the text prompt. ~\cite{wu2022membership} is the first to analyze the membership leakage problem in text-to-image generation models, where whether a certain image is used to train the target text-to-image model is inferred. Specifically, ~\cite{wu2022membership} proposes three intuitions on the membership information and four attack methods accordingly. Experiments show that all the proposed attack methods achieve impressive results, highlighting the threat of membership leakage.

\begin{figure*}[!tbp]\centering
\includegraphics[width=1.0\linewidth]{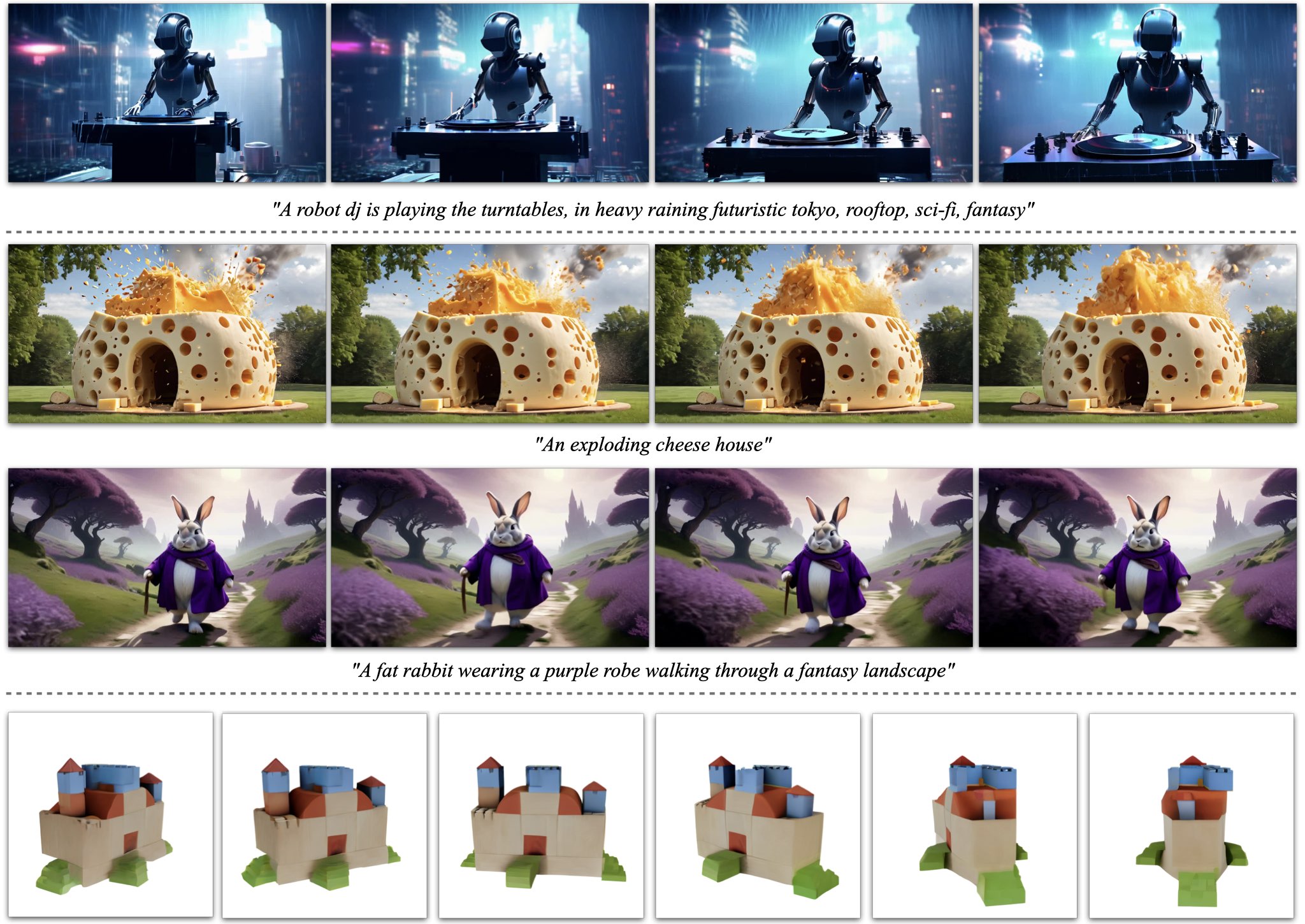}\\
\caption{Text-to-video generation by Stable Video Diffusion~\cite{blattmann2023stable} from Stable AI.}
\label{fig:stable_video_diffusion}
\end{figure*}

\section{Applications beyond image generation}
\label{sec:applications}

The advancement of diffusion models has inspired various applications beyond image generation, such as text-to-X where X refers to a modality such as video, and text-guided image editing. We introduce pioneering work as follows.

\subsection{Text-to-X generation}
\subsubsection{Text-to-art}

Artistic painting is an interesting and imaginative area that benefits from the success of generative models. Despite the progress of GAN-based painting\cite{jabbar2021survey}, they suffer from the unstable training and model collapse problem brought by GAN. Recently, multiple works have presented impressive images of paintings based on diffusion models, investigating improved prompts and different scenes. 
Multimodal guided artwork diffusion (MGAD) ~\cite{huang2022draw} refines the generative process of diffusion model with multimodal guidance (text and image) and achieves excellent results regarding both the diversity and quality of generated digital artworks.  In order to maintain the global content of the input image, DiffStyler~\cite{huang2024diffstyler} proposes a controllable dual diffusion model with learnable noise in the diffusion process of the content image. During inference, explicit content and abstract aesthetics can both be learned with two diffusion models. Experimental results show that DiffStyler~\cite{huang2024diffstyler} achieve excellent results on both quantitative metrics and manual evaluation. To improve the creativity of Stable Diffusion model,  ~\cite{wu2022creative} proposes two directions of textual condition extension and model retraining with the Wikiart dataset, enabling the users to ask the famous artists to draw novel images. ~\cite{gallego2022personalizing} personalizes text-to-image generation by customizing the aesthetic styles with a set of images, while ~\cite{jain2022vectorfusion} extends generated images to Scalable Vector Graphics (SVGs) for digital icons or arts.  In order to improve computation efficiency, ~\cite{rombach2022text} proposes to generate artistic images based on retrieval-augmented diffusion models. By retrieving neighbors from specialized datasets (e.g., Wikiart), ~\cite{rombach2022text} obtains fine-grained control of the image style. In order to specify more fine-grained style features (e.g., color distribution and brush strokes),  ~\cite{pan2023arbitrary} proposes supervised style guidance and self-style guidance method, which can generate images of more diverse styles.

\subsubsection{Text-to-video}
\textbf{Early studies.}  Since video is just a sequence of images, a natural application of text-to-image is to make a video conditioned on the text input. Conceptually, text-to-video DM lies in the intersection between text-to-image DM and video DM.   Make-A-Video~\cite{singer2022make} adopts a pretrained text-to-image DM to text-to-video and Video Imagen~\cite{ho2022imagen} extends an existing video DM method to text-to-video.  Other representative text-to-video diffusion models include ModelScope~\cite{wang2023modelscope}, Tune-A-Video~\cite{wu2023tune}, and VideoCrafter~\cite{chen2023videocrafter1,chen2024videocrafter2,xing2023dynamicrafter}. The success of text-to-video naturally inspires a future direction of movie generation based on text inputs. Different from general text-to-video tasks, story visualization requires the model to \textit{reason} at each frame about whether to maintain the consistency of actors and backgrounds between frames or scenes, based on the story progress ~\cite{rahman2022make}. Make-A-Story~\cite{rahman2022make}  uses an autoregressive diffusion-based framework and visual memory module to maintain consistency of actors and backgrounds across frames, while AR-LDM~\cite{pan2022synthesizing} leverages image-caption history for coherent frame generation.  Moreover, AR-LDM~\cite{pan2022synthesizing} shows the consistency for unseen characters, and also the ability for real-world story synthesis on a newly introduced dataset VIST~\cite{huang2016visual}.

\begin{figure*}[!tbp]\centering
\includegraphics[width=1.0\linewidth]{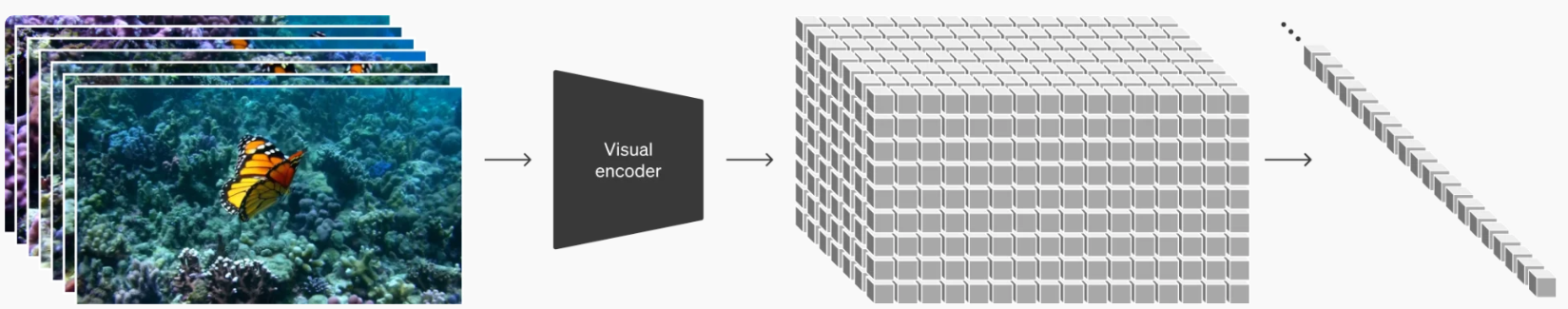}\\
\caption{Sora~\cite{openai2024video} from Open AI. Sora represents video frames by compressing them to image patches with a transformer-base backbone.} 
\label{fig:sora}
\end{figure*}
\textbf{Recent process.} More recently, Stable Video Diffusion~\cite{blattmann2023stable} from Stable AI achieves significant performance improvements for text-to-video and image-to-video generation. It identifies and evaluates different training stages of applying diffusion models to video synthesis, and introduces a systematic training process including the captioning and filtering strategies.  OpenAI launches the state-of-the-art video generation model  Sora~\cite{openai2024video}, which is capable of generating a minute of high-fidelity video.  Inspired by large language models that turn different data (like text and code) into tokens,   Sora~\cite{openai2024video}  first unifies diverse types of videos and images as patches and compresses them to a lower-dimensional latent space, as shown in Figure~\ref{fig:sora}. Sora then decoposes the representations into spacetime patches and performs the diffusion process based on the transformer backbone. As Sora~\cite{openai2024video} is not open-sourced yet, some studies aim to provide open access to advanced video generation models, such as  Open-Sora~\cite{opensora} and  Open-Sora-Plan~\cite{pku_yuan_lab_and_tuzhan_ai_etc_2024_10948109}.

\subsubsection{Text-to-3D}

3D object generation is evidently much more sophisticated than 2D image synthesis task. DeepFusion~\cite{poole2022dreamfusion} is the first work that successfully applies diffusText-guided creative generation models to 3D object synthesis. Inspired by Dream Fields~\cite{jain2022zero} which applies 2D image-text models (i.e., CLIP) for 3D synthesis, DeepFusion~\cite{poole2022dreamfusion} trains a randomly initialized NeRF~\cite{mildenhall2021nerf} with the distillation  of  a pretrained 2D diffusion model (i.e., Imagen). However, according to  Magic3D ~\cite{lin2023magic3d},  the low-resolution image supervision and extremely slow optimization of NeRF result in low-quality generation and long processing time of DeepFusion~\cite{poole2022dreamfusion}.  For higher-resolution results, Magic3D ~\cite{lin2023magic3d} proposes a coarse-to-fine optimization approach with coarse representation as initialization as the first step, and optimizing mesh representations with high-resolution diffusion priors.  Magic3D ~\cite{lin2023magic3d} also accelerates the generation process with a sparse 3D hash grid structure. 3DDesigner~\cite{li20223ddesigner}  focuses on another topic of 3D object generation, \textit{consistency}, which indicates the cross-view correspondence. With low-resolution results from NeRF-based condition module as the prior, a two-stream asynchronous diffusion module further enhances the consistency and achieves 360-degree consistent results. Apart from 3D object generation from text, recent work Zero-1-to-3~\cite{liu2023zero1to3} has achieved great attention by enabling zero-shot novel view synthesis and 3D reconstruction from a single image, inspiring various follow-up studies such as Vivid-1-to-3~\cite{kwak2024vivid}.

\subsection{Text-guided image editing}

Diffusion models not only significantly improve the quality of text-to-image synthesis, but also enhance text-guided image editing.  Before DM gained popularity, zero-shot image editing had been dominated by GAN inversion methods~\cite{abdal2020image2stylegan++,bau2020semantic,richardson2021encoding,tov2021designing,pernuvs2025fice,dere2024conditional} combined with CLIP. However, GAN is often constrained to have limited inversion capability, causing unintended changes to the image content. In this section, we discuss pioneering studies for image editing based on diffusion models.

\textbf{Inversion for image editing.} A branch of studies edits the images by modifying the noisy signals in the diffusion process.  SDEdit\cite{meng2022sdedit} is a pioneering work that edits images by iteratively denoising through a stochastic differential equation (SDE).  Without any task-specific training, SDEdit\cite{meng2022sdedit} first add noises to the input (such as stroke painting), then subsequently denoises the noisy image through the SDE prior to increase image realism. DiffusionCLIP~\cite{Kim_2022_CVPR} further adds text control in the editing process by fine-tuning the diffusion model at the reverse DDIM~\cite{song2020denoising}  process with CLIP-based loss.  
Due to the local linearization assumptions, DDIM may lead to incorrect image reconstruction with the error propagation ~\cite{wallace2023edict}. To mitigate this problem, Exact Diffusion Inversion via Coupled Transformations (EDICT)~\cite{wallace2023edict} proposes to maintain two coupled noise vectors in the diffusion process and achieves higher reconstruction quality than DDIM~\cite{song2020denoising}. Another work ~\cite{mokady2023null} introduces an accurate inversion technique for text-guided editing by pivotal inversion and null-text  optimization, showing high-fidelity editing of various real images.  To improve the editing efficiency, LEDITS++~\cite{brack2024ledits++} proposes a novel inversion approach without tuning and optimization, which could produce high-fidelity results with a few diffusion steps.

\begin{figure}[!tbp]\centering
\includegraphics[width=0.7\linewidth]{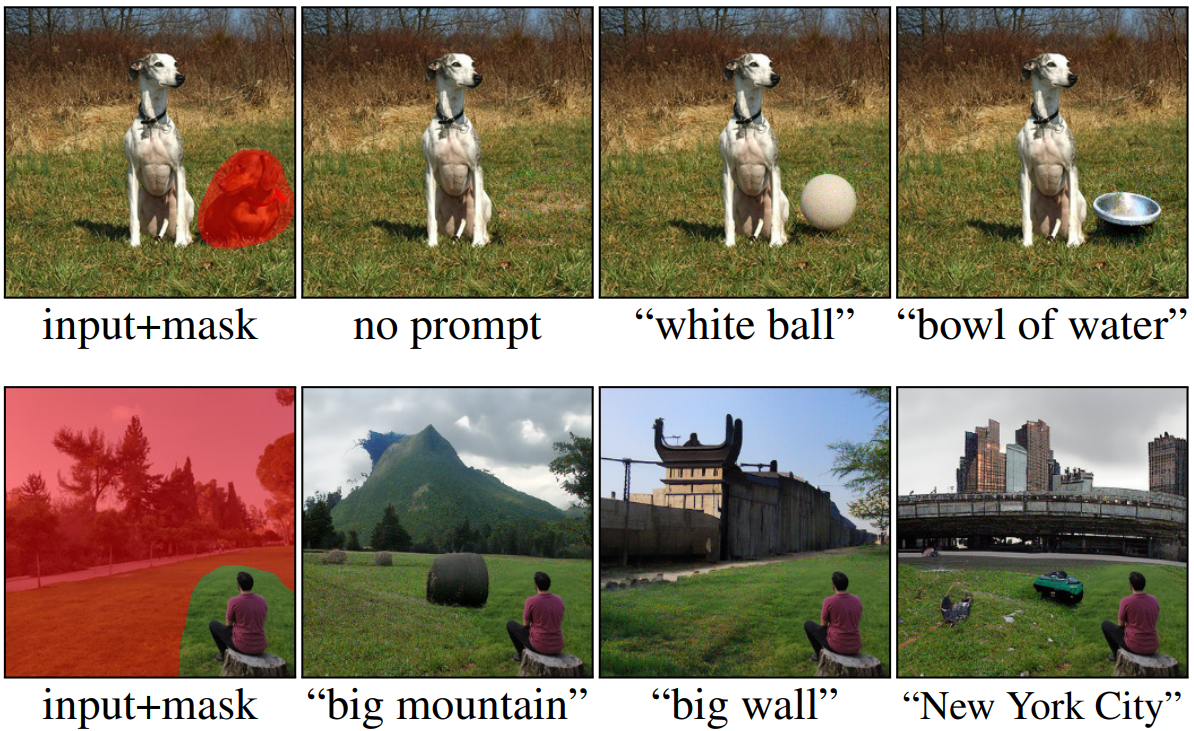}\\
\caption{Edit images by mask control in Blended Diffusion~\cite{avrahami2022blended1}.}
\label{fig:blend_diffusion}
\end{figure}

\textbf{Editing with mask control.} A branch of work manipulates the image mainly on a local (masked) region~\cite{avrahami2022blended1}, as shown in Figure~\ref{fig:blend_diffusion}. The difficulty lies in guaranteeing a seamless coherence between the masked region and the background. 
To guarantee the seamless coherence between the edited region and the remaining part, Blended diffusion~\cite{avrahami2022blended1} spatially blends noisy image with the local text-guided diffusion latent in a progressive manner. This approached is further improved for  a blended latent diffusion model in ~\cite{avrahami2023blended} and a multi-stage variant in ~\cite{ackermann2022high}. Different from ~\cite{avrahami2022blended1,avrahami2023blended,ackermann2022high} that requires a manually designed mask,  DiffEdit~\cite{couairon2022diffedit} proposes to automatically generate the mask to indicate which part to be edited. 

\textbf{Expanded editing with flexible texts.} Some studies enable more types of image editing with flexible text inputs. Imagic~\cite{kawar2022imagic} is the first to perform text-based semantic edits to a single image, such as postures or composition of multiple objects. Specifically, Imagic~\cite{kawar2022imagic} first obtains an optimized embedding for the target text, then linearly interpolates between the target text embedding and the optimized one. This generated representation is then sent to the fine-tuned model and generates the edited images.  To solve the problem that a simple modification of text prompt may leads to a different output, Prompt-to-Prompt~\cite{hertz2022prompt}  proposes to use a cross-attention map during the diffusion progress, which represents the relation between each image pixel and word in the text prompt. InstructPix2Pix~\cite{brooks2023instructpix2pix} works on the task of editing the image with human-written instructions. Based on a large model (GPT-3) and a text-to-image model (Stable diffusion), ~\cite{brooks2023instructpix2pix} first generates a dataset for this new task, and trains a conditional diffusion model  InstructPix2Pix which generalizes to real images well. However, it is admitted in ~\cite{brooks2023instructpix2pix} there still remain some limitations, e.g., the visual quality of generated datasets limits the model.

There are also other interesting tasks regarding image editing. Paint by example~\cite{yang2022paint} proposes a  semantic image composition problem reference image is semantically transformed and harmonized before blending into another image~\cite{yang2022paint}. MagicMix~\cite{liew2022magicmix}  proposes a new task called semantic mixing, which blends two different semantics (e.g.,corgi and coffee machine) to create a new concept (corgi-like coffee machine). SEGA~\cite{brack2023sega} allows for subtle and extensive image edits by instructing the diffusion models with semantic control. Specifically,  it interacts with the diffusion process to flexibly steer it along semantic directions.

\begin{figure}[!tbp]\centering
\includegraphics[width=1.0\linewidth]{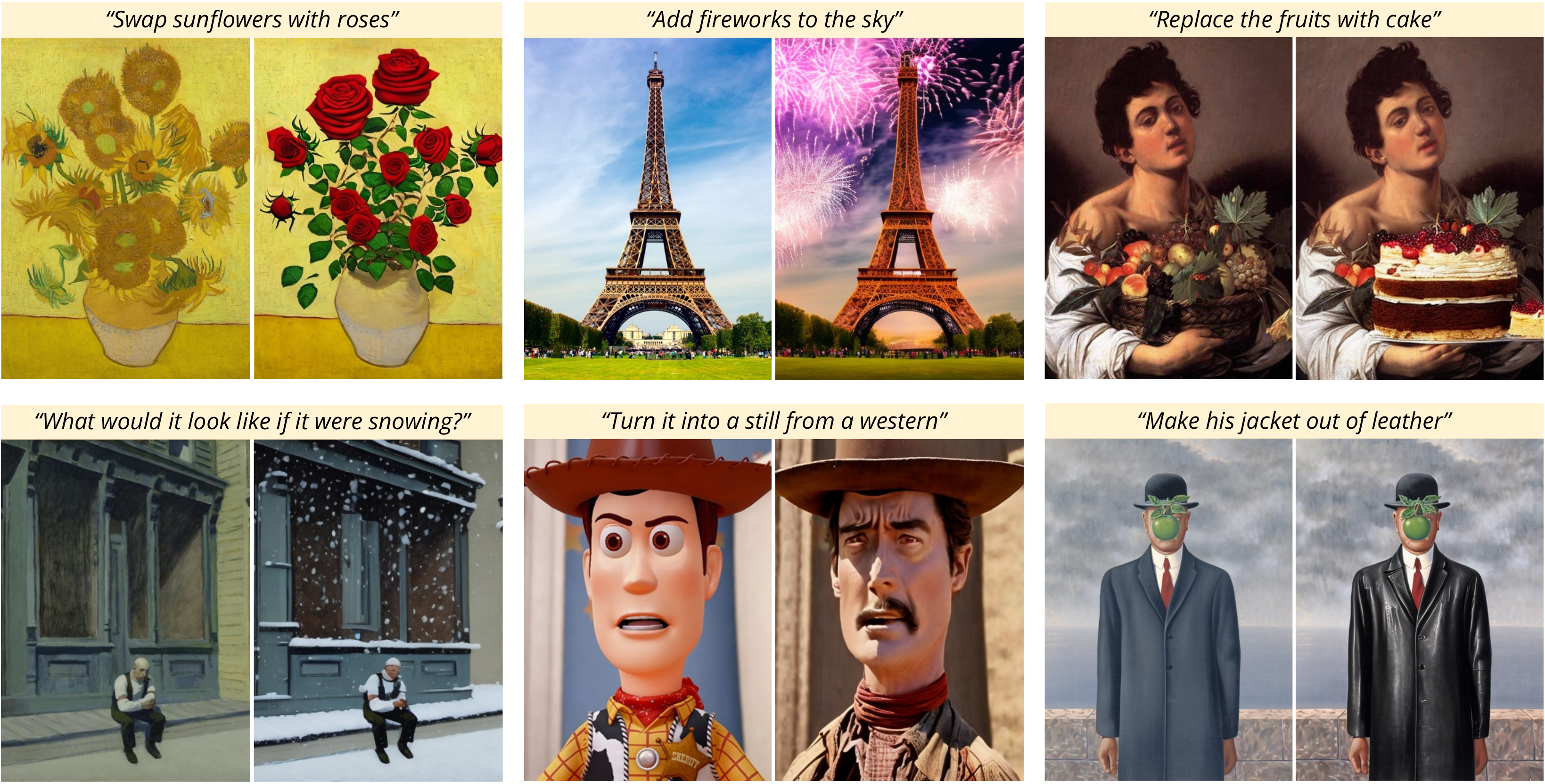}\\
\caption{Image editing by InstructPix2Pix~\cite{brooks2023instructpix2pix}. InstructPix2Pix~\cite{brooks2023instructpix2pix} allows the users to edit an existing image by simply giving textural instructions, such as ``Swap sunflowers with roses".}
\label{fig:instructpix2pix}
\end{figure}

\section{Challenges and outlook}

\subsection{Challenges}

\textbf{Challenges on ethical issues and dataset bias.} Text-to-image models trained on large-scale unfiltered data may suffer from even reinforce the biases from the training dataset,  leading to the generation of inappropriate (e.g., offensive, insulting, or
threatening information) ~\cite{schramowski2023safe}  or unfair content regarding social groups~\cite{struppek2022biased,bansal2022well}.  Moreover, the current models predominantly adopt English as the default language for input text. This might further put those people who do not understand English in an unfavored situation~\cite{friedrich2024multilingual,struppek2023exploiting}. 

\textbf{Challenges on security risks.} As the diffusion models improve, it is becoming more 
 difficult to detect generated images from the real ones. This brings security risks since the generated images may be used for malicious purposes like falsifying electronic evidence~\cite{sha2023fake}. Moreover, the diffusion models suffer from security risks, such as backdoor attack~\cite{struppek2022rickrolling} and privacy issues~\cite{wu2022membership}.

\textbf{Challenges on data and computation.} As widely recognized, the success of deep learning heavily depends on the scaled training data and huge computation resources. In the context of text-to-image DM, this is especially true. For example, the major frameworks, such as Stable Diffusion~\cite{rombach2022high} and DALL-E 2~\cite{ramesh2022hierarchical}, are all trained with hundreds of millions of image-text pairs. Moreover, the computation overhead is so large that it renders the opportunities to train such a model from scratch to large companies, such as OpenAI~\cite{ramesh2022hierarchical} and Google~\cite{saharia2022photorealistic}. Despite the advancements to accelerate the training and inference of diffusion models, it is still challenging to effectively adopt the diffusion models in efficiency-oriented environments, such as edge devices.

\subsection{Outlook}
 
\textbf{Safe and fair applications.} With the wide application of text-to-image models, how to mitigate the ethical issues and security risks of current text-to-image models is demanding and challenging. Possible directions include a more diverse and balanced dataset to mitigate issues like  race and gender, advanced methods for detecting generated images, and robust diffusion models against various attacks.

\textbf{Unified multi-modality framework.}  Text-to-image generation can be seen as  part of the multi-modality learning. Most works focus on the single task of  text-to-image generation, but unifying multiple task into a single model can be a promising trend. For example, UniD3~\cite{hu2022unified}  unify text-to-image generation and image captioning with a single diffusion model. The unified multi-modality model can boost each task by learning  representations from each modality better, and may bring more inspirations of how model understands the multi-modality data. 

\textbf{Collaboration with other fields.} In the past few years, deep learning has made great progress in multiple areas, such as multi-modal GPT-4~\cite{openai2024gpt4}. Prior studies have investigated how to collaborate diffusion models with models from other fields, such as the recent work that deconstructs a diffusion model to an autoencoder~\cite{chen2024deconstructing} and the adoption of GPT-3~\cite{openai2024gpt3} in InstructPix2Pix~\cite{brooks2023instructpix2pix}. There are also studies applying diffusion models in vision applications, such as image restoration~\cite{liu2024adaptbir}, depth estimation~\cite{kim2024depth,xu2024two},  image enhancement~\cite{panagiotou2024denoising} and classification~\cite{li2023your}. Further collaborations between text-to-image diffusion model and recent findings in active research fields are an exciting topic to be explored.

 \bibliographystyle{elsarticle-num-names} 
 \bibliography{bib_short}

\end{document}